\ifdefined\pdfminorversion\pdfminorversion=7\fi 
\PassOptionsToPackage{hyphens}{url} 
\documentclass[11pt, a4paper, copyright, gr]{google}
\usepackage[authoryear, sort&compress, round]{natbib}
\usepackage{array}
\newcolumntype{C}[1]{>{\centering\arraybackslash}p{#1}}
\usepackage{longtable}
\newcommand{\rowrule}{\arrayrulecolor{black!25}\specialrule{0.3pt}{2pt}{2pt}\arrayrulecolor{black}}


\usepackage{listings}
\usepackage{upquote}

\usepackage{algpseudocode}
\newcommand{\dothrule}{\noindent\makebox[\linewidth]{\cleaders\hbox to 0.4em{\hss$\cdot$\hss}\hfill}}
\newcommand{\blank}{\rule[-0.2ex]{1.5em}{0.9pt}}
\definecolor{stepcolor}{RGB}{34, 139, 34}
\uselogo{}

\title{Google Earth AI Geospatial Reasoning Agent: Capabilities and a Systematic Evaluation}

\correspondingauthor{noamslonim@google.com}
\reportnumber{} 

\renewcommand{\today}{October 11, 2026}

\author[1,*]{Noam Slonim}
\author[1,*]{Aviv Slobodkin}
\author[1,*]{Gil Fidel}
\author[1,*]{Gal Moshitch}
\author[1,*]{Mimi Sun}
\author[1,*]{Lily Mihalkova}
\author[1,*]{Joydeep Paul}
\author[1,*]{Per Bjornsson}
\author[2,*]{Sydney Lin}
\author[1,*]{Jamie McPike}
\author[3]{Samapriya Roy}
\author[2]{Utkarsh Patel}
\author[2]{Gia Jung}
\author[1]{Zachary Luna}
\author[1]{Allison Berman}
\author[1]{William Cai}
\author[1]{Xiao Wu}
\author[1]{Mohit Agarwal}
\author[1]{Mandar Sharma}
\author[1]{Aliza Hoffman}
\author[1]{Mark Young}
\author[1]{Rama Pasumarthi}
\author[1]{Roy Lee}
\author[1]{Ruth Alcantara}
\author[1]{Luke Barrington}
\author[1]{Aaron Bell}
\author[1]{Reuven Sayag}
\author[1]{Avinatan Hassidim}
\author[1]{Yossi Matias}
\author[1]{Tyler McHenry}
\author[1]{Ronald Ho}
\author[1]{Christopher Haire}
\author[2]{Oliver Guinan}
\author[1]{Niv Efron}
\author[1,$\dagger$]{Tim Thelin}
\author[1,$\dagger$]{Gautam Prasad}
\author[1,$\dagger$]{Shravya Shetty}

\affil[1]{Google Research}
\affil[2]{Google Geo}
\affil[3]{Spatial Bytes}
\affil[*]{Core contributor}
\affil[$\dagger$]{Co-last author}

\begin{abstract}
Google Earth AI is a family of geospatial AI models, together with an agent that orchestrates them. We present this agent, Google Earth AI's Geospatial Reasoning agent (GRA), a Gemini-powered research system that brings together a broad suite of geospatial data sources, foundation models, and unique tools to answer questions about our planet, from factual lookups to complex multi-step analysis and prediction. GRA organizes this capability suite into modular domain skills and pairs them with upfront geospatial planning to decompose complex queries into multi-stage analytical workflows. Evaluating such systems is challenging because real-world geospatial problems admit multiple valid workflows, require discovering data across large catalogs, and rely on evolving physical ground truth. To address this, we introduce two benchmarks and release their development splits. GeoExpertQueries poses complex analytical problems grounded in real-world practitioner use cases and refined by geospatial experts; each problem is paired with a consensus golden workflow derived from three independent expert workflows, and responses are scored against it by a human-validated, tool-agnostic autorater. GeoKTools consists of templated queries with deterministic reference solvers, in which the number of required tool types is varied systematically. We complement our evaluation with two external benchmarks with deterministic answers, CBGB and UnivEARTH, the latter of which we find to be saturated. To isolate the specific contribution of GRA's tools and its agentic orchestration, we compare four configurations, each run with both Gemini 3.1 Pro and Gemini 3.8 Flash: a standalone LLM, the LLM with basic tools, the LLM with GRA's skills and tools but without its upfront planning and execution guidance, and GRA itself. With at most basic tools (search, maps, offline code execution), an advanced model recovers a substantial fraction of the answer on templated queries but satisfies only a small fraction of the expert workflows on open-ended problems; GRA performs well on both. The largest step comes from GRA's geospatial skills and tools, which on their own lift performance far above the baselines; on the open-ended problems, GRA's upfront planning and execution guidance then add a substantial further gain with Gemini 3.8 Flash, allowing the more recent Flash model to match or exceed Pro at lower cost. Our methodology and findings provide a rigorous framework for assessing open-ended geospatial agents and the first broad quantitative evaluation of GRA on challenging, real-world geospatial problems.

\end{abstract}

\begin{document}

\maketitle

\section{Introduction}\label{sec:intro}

A growing range of critical decisions depend on integrating heterogeneous geospatial data: satellite imagery, demographic statistics, weather forecasts, and environmental databases. Assessing flood risk, monitoring deforestation, or siting renewable energy infrastructure each demands multi-source analysis that has traditionally required deep domain knowledge and manual data integration. Large language models (LLMs) have enabled a new class of \emph{geospatial reasoning agents} that accept queries in natural language and autonomously orchestrate domain-specific tools to produce actionable insights. A leading example is Google Earth AI's Geospatial Reasoning agent (GRA), which orchestrates the Google Earth AI family of geospatial AI models \citep{earthai2025}. Powered by Gemini, GRA combines general-purpose LLM reasoning with a broad suite of geospatial assets, many of them unique to Google, including foundation models for remote sensing, population dynamics, and satellite embeddings; weather and flood forecasting; Google Maps Data; and planetary-scale catalogs such as Google Earth Engine and Data Commons (Section~\ref{sec:eaia}). Together, these enable users without domain expertise to conduct complex geospatial analyses: GRA finds the relevant sources, combines data such as map layers, Earth observation products, and demographic statistics, and carries out intermediate computation, so that a question posed in plain language returns an analysis rather than a dataset.

The current GRA builds on the system introduced in \citet{earthai2025} to support tasks ranging from factual question answering to multi-step spatial analysis, prediction, and interactive map visualization. Two design choices let the agent use this broad capability suite effectively. First, capabilities are packaged as modular skills. Their tool definitions stay hidden until the agent loads them, so context is spent only on what a query needs, and new capabilities can be added without making existing ones harder to use. Large spatial datasets are likewise kept out of the context and passed between tools by reference as scratch datasets. Second, GRA begins each query with an upfront planning stage that reasons over documented capabilities and their limitations to lay out a retrieval and processing plan before any data tool is called. It then executes against that plan, with guardrails for dead ends, substituting proxy datasets when an ideal measurement is unavailable (Section~\ref{sec:systems}).

Evaluating reasoning agents rigorously is a recognized challenge, both in geospatial AI and in other agentic domains such as software engineering and scientific discovery \citep{jimenez2024swebench, chen2025scienceagentbench}. Four factors make it particularly difficult. First, complex geospatial problems rarely have a single correct answer: outcomes depend on the data sources consulted, the spatial and temporal boundaries set, and the analytical methods applied. Second, multiple valid workflows may lead to equally sound results (e.g., a flood risk assessment based on official flood-zone maps or on high-resolution flood simulations), making it difficult to define a single reference process. Third, a geospatial query typically does not include the data needed to answer it, which the agent must find in large, heterogeneous catalogs. Data discovery thus becomes part of what is evaluated and, because systems differ in the catalogs and tools they can reach, a poor result may reflect limited data access rather than poor reasoning, which makes systems hard to compare. Fourth, ground truth in this domain is not stable: satellite archives are reprocessed, reference datasets are revised, and administrative or protected-area boundaries are redrawn, so an answer that was correct when a benchmark was built may no longer be correct when the benchmark is run.

Faced with these challenges, existing geospatial benchmarks have typically constrained the problem: they fix the tool set and the data catalog in advance, so that a query has one correct answer, one reference workflow, and nothing to discover. Scoring then reduces to checking whether the system called the right functions with the right parameters. This is practical, but it removes the open-endedness of real analytical work and ties any conclusion to one particular tool set. We therefore evaluate on multiple benchmarks with complementary evaluation paradigms, including benchmarks that are tool-agnostic by design.

It is also hard to measure what geospatial tools and their orchestration add, since a frontier LLM already knows a great deal about the world and can answer some questions without them. Our original geospatial reasoning preprint \citep{earthai2025} evaluated the system on 100 factual questions and 10 crisis-response prompts scored against hand-written rubrics rather than reference solutions, and noted the difficulty of establishing whether the agent had genuinely carried out a geospatial analysis or had simply synthesized information already available on the open web.

Here, we present a systematic evaluation of GRA, designed to quantify what each layer of the system contributes. To our knowledge, no prior study of geospatial agents has measured this, since doing so requires varying the tools and orchestration around a fixed model rather than only the model. We compare four configurations built on the same Gemini model: a standalone LLM, the LLM with basic tools, the LLM with GRA's geospatial skills and tools in a reactive loop, and the full GRA. This separates what the domain tools contribute from what upfront planning and execution guidance add. We run the comparison with both Gemini 3.1 Pro and Gemini 3.8 Flash, on two new benchmarks, GeoExpertQueries and GeoKTools, and two adapted external ones, CBGB \citep{cardille2025cbgb} and UnivEARTH \citep{kao2026univearth}. A companion study \citep{ma2026} presents the Planetary Prediction Engine (PPE), which uses Earth AI tools for autonomous geospatial prediction, such as spatial regression, downscaling, and epidemiological nowcasting; the present study focuses on open-ended analytical queries.

\begin{figure}[t]
\centering
\includegraphics[width=\textwidth, trim=30 22 30 28, clip]{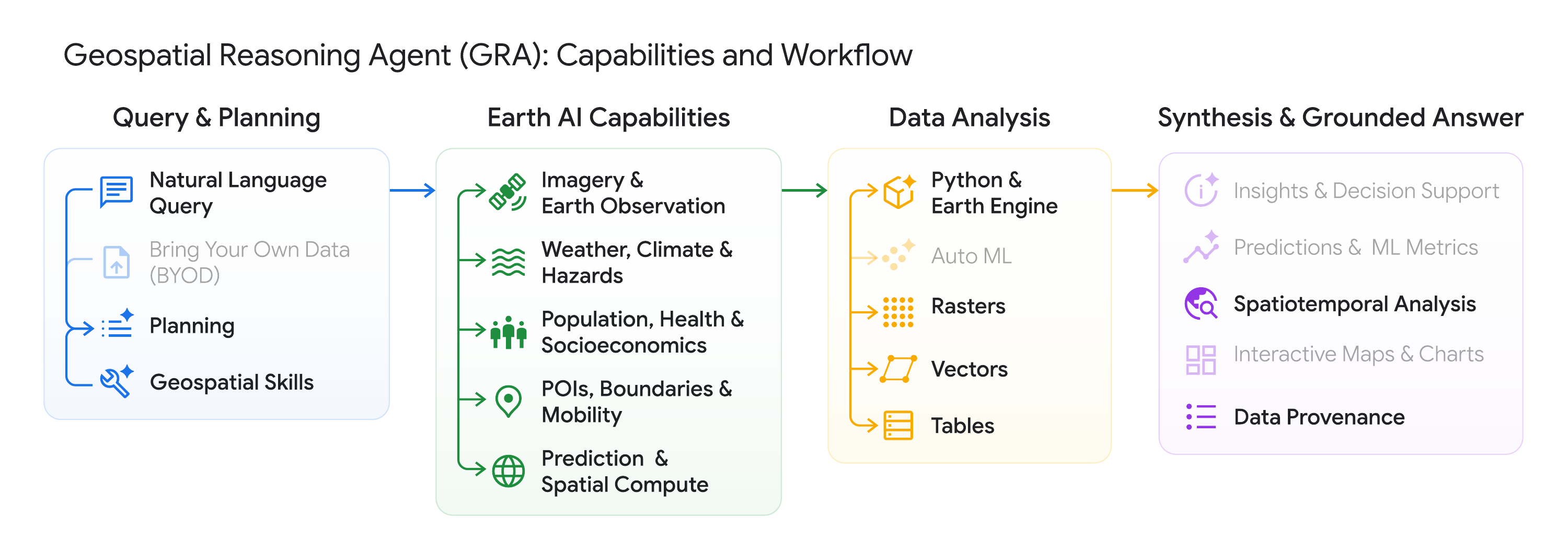}\\[3mm]
\includegraphics[width=\textwidth, trim=30 21 30 29, clip]{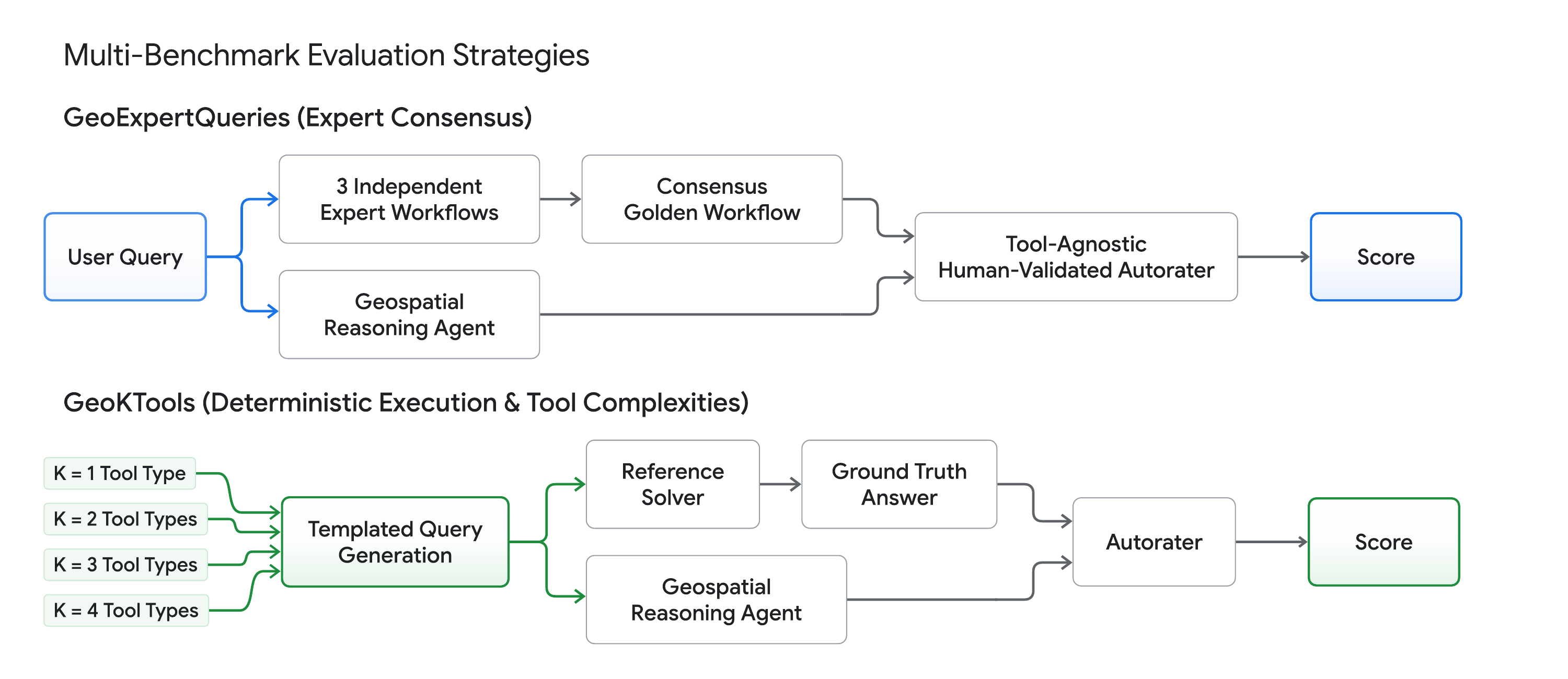}
\caption{The Geospatial Reasoning agent (GRA) and how we evaluate it. The agent plans its approach to a natural-language query, draws on geospatial capabilities across five pillars (Table~\ref{tab:eaia-pillars}), analyzes the retrieved data, and synthesizes a grounded answer. The bottom panel shows how our two benchmarks score the agent's responses. Greyed-out items are not evaluated in this report.}
\label{fig:overview}
\end{figure}

Our study makes the following contributions:

\begin{itemize}
  \item \textbf{GRA: bringing geospatial capabilities together with modular skills and upfront planning.} We present the expanded Geospatial Reasoning agent (GRA), which brings a wide range of geospatial capabilities together behind a single natural-language interface, drawing on Google Earth AI models such as the Remote Sensing Foundations models (RSFM), Population Dynamics Foundation Model (PDFM), AlphaEarth Foundations, WeatherNext, and operational flood forecasting, as well as Google Maps Data, Google Earth Engine, and Data Commons. To scale across this breadth without overwhelming model context, GRA packages domain capabilities as modular skills loaded on demand, keeps large datasets outside the model's context as scratch datasets, and pairs both with capability-aware upfront geospatial planning (Figure~\ref{fig:overview}; Section~\ref{sec:systems}).
  \item \textbf{A multi-benchmark evaluation suite and a controlled four-configuration comparison.} We evaluate on four complementary benchmarks. GeoExpertQueries comprises 101 open-ended problems (76 held out for testing), grounded in real-world practitioner use cases and refined by geospatial domain experts across areas such as crisis resilience, public health, and site selection, each paired with a consensus golden workflow (from three independent experts blind to GRA's tools) and scored by a human-validated, tool-agnostic autorater that credits a sound analysis without requiring a prescribed sequence of tool calls. GeoKTools comprises 305 templated queries with deterministic answers, 220 of them held out for testing, drawn from 92 expert-written templates spanning 24 tool types, in which the number of required tool types is varied systematically ($K=1,2,3,4$). We complement these with two externally developed benchmarks, CBGB \citep{cardille2025cbgb} and UnivEARTH \citep{kao2026univearth}, which provide an external check on our internally designed ones. Comparing the four configurations on the same underlying model across this suite isolates the contribution of specialized geospatial capabilities from that of upfront planning and execution guidance (Section~\ref{sec:systems}). We release the development splits for both new benchmarks, together with their autorater prompts and scoring code, and, for GeoKTools, a specification of each template's reference solver (Section~\ref{sec:benchmarks}).
  \item \textbf{Key findings.} GRA substantially outperforms the two simpler baselines, a standalone LLM and the same LLM with basic tools, on GeoExpertQueries, GeoKTools and CBGB. On GeoExpertQueries it scores 63\footnote{All scores in this paper are reported on a 0--100 scale, but each benchmark has its own scoring scheme, as explained in Section~\ref{sec:benchmarks}. Scores are therefore meaningful for comparing systems within a benchmark, not across benchmarks.}, while both baselines prove effectively unusable, satisfying only a small fraction of the requirements set out in the expert workflows. On GeoKTools it scores 86 against 41 for the strongest baseline, a standalone LLM with no tools at all (Section~\ref{sec:res-geoktools}), and on CBGB it scores above 80 while neither baseline exceeds 20. The largest gain comes from GRA's geospatial skills and tools, which on their own lift performance far above both baselines, for example from 3 to 52 on GeoExpertQueries with Gemini 3.8 Flash. GRA's upfront planning and execution guidance then help the agent make fuller use of those tools on the hardest, open-ended queries, raising GeoExpertQueries from 52 to 63 with Gemini 3.8 Flash. The one exception is UnivEARTH, where every configuration, including a standalone LLM with no Earth Engine access, scores above 85, indicating that this benchmark has become saturated as models answer directly from published source text without performing geospatial analysis. Within GRA, the more recent Gemini 3.8 Flash outperforms Gemini 3.1 Pro on GeoExpertQueries and GeoKTools at lower cost, and matches it on CBGB.
\end{itemize}

\section{Related Work}\label{sec:related}

The vision of AI-powered geospatial analysis has evolved rapidly, from early proposals for autonomous GIS \citep{li2023autonomous, zhang2023geogpt, akinboyewa2025giscopilot} to full-stack agentic systems that combine LLMs with domain-specific tools. Recent systems include ClimateAgent \citep{kim2025climateagent}, NORA \citep{zhou2026nora}, Earth-Agent \citep{feng2026earthagent}, and GeoNatureAgent \citep{diaz2026geonatureagent}, each pairing an LLM with specialized geospatial tools and, in some cases, multi-agent orchestration. However, prior academic frameworks typically rely on public remote-sensing catalogs and open-source GIS libraries, whereas GRA further integrates Google's own foundation models, forecasting systems, Google Maps Data, and other unique assets. Moreover, evaluations of prior systems have typically compared the full system against a standalone LLM baseline \citep{zhou2026nora} or varied the backbone LLM while holding the tool set constant \citep{feng2026earthagent, diaz2026geonatureagent}. To our knowledge, no prior work has varied the capability layer itself, comparing a standalone model, the same model with general-purpose tools, and the same model with domain-specific geospatial tools, with and without an orchestration layer built around them.

Agent design more broadly has also shifted. Early frameworks often split work across multiple specialized agents that coordinate through conversation \citep{wu2024autogen}, but making such systems effective is not trivial and can introduce failure modes of their own \citep{cemri2025multiagent}. Thus, practitioner guidance increasingly favors simple, composable designs \citep{anthropic2024agents}. One approach is to package domain capabilities as skills that a single agent loads on demand, keeping only short descriptions in context until a skill is needed \citep{anthropic2025skills}. In parallel, planning-first designs, in which the agent lays out its steps before invoking tools, have been proposed as an alternative to interleaving reasoning and action step by step \citep{yao2023react}, with reported gains in accuracy and efficiency \citep{xu2023rewoo, kim2024llmcompiler}. Both matter especially for geospatial reasoning. First, petabyte-scale catalogs and dozens of specialized tools cannot all be exposed at once in context. Second, long multi-step analyses can benefit from a plan laid out in advance. GRA combines both designs, and our evaluation shows the value of this combined planning and orchestration layer by comparing the full GRA with a configuration that uses the same model, skills, and tools, but without this layer (Section~\ref{sec:systems}).

In parallel to GRA, commercial platforms for AI-powered geospatial analysis have emerged, including Agentic Earth\footnote{\url{https://agenticearth.io}}, EmergenCity AI\footnote{\url{https://emergencity.ai}}, and Planet Agentic AI\footnote{\url{https://ai.planet.com/}}. However, none of these has been accompanied by published evaluation results, making systematic comparison infeasible at this time.

A growing number of benchmarks evaluate AI systems on geospatial problems. Next, we briefly discuss these benchmarks, grouped by the evaluation paradigm. Code generation benchmarks such as CBGB \citep{cardille2025cbgb} and UnivEARTH \citep{kao2026univearth} test whether LLMs can write executable geospatial code for Google Earth Engine, and compare numerical or yes/no outputs against reference answers. Tool-calling benchmarks such as GeoBenchX \citep{krechetova2025geobenchx} and GeoAgentBench \citep{yu2026geoagentbench} evaluate whether agents can select and invoke the right GIS functions with correct parameters, but each defines its own tool catalog, tying evaluation to those specific tools. Several benchmarks \citep{shabbir2026thinkgeo, xu2026earthspatial, jiang2025geogrid, feng2026earthagent} require agents to process satellite or gridded imagery provided as input, testing visual perception and spatial reasoning rather than the data-discovery and analytical orchestration that is the focus of our study. DisasterBench \citep{chen2026disasterbench} evaluates static workflow planning for disaster response, but plans are scored without execution. AtmosSci-Bench \citep{li2025atmossci} tests domain knowledge via graduate-level atmospheric science questions without tool use, notably finding that domain-specific fine-tuned models underperform general reasoning models, which suggests that domain specialization at the model level is not where the advantage lies. Most of these benchmarks test a single capability in isolation (code generation, tool calling, visual perception, or planning) rather than the end-to-end, multi-source workflows that characterize complex real-world geospatial problems. None combines open-ended practitioner tasks, independently authored expert reference workflows, tool-agnostic scoring, and a comparison of configurations built on a fixed model. This is the gap our evaluation suite is designed to fill.

We develop two benchmarks. GeoExpertQueries poses open-ended problems grounded in practitioner use cases and scores responses against a consensus of independently authored expert workflows, using tool-agnostic scoring. GeoKTools provides controlled queries that each require a known number of distinct tool types, so that performance can be measured as this number grows. To provide an additional external check, we also adopt two benchmarks from the landscape above, selected because their answers are unambiguous, they were developed independently of our system, and they test complementary capabilities. CBGB \citep{cardille2025cbgb} poses 45 expert-curated geospatial questions across three difficulty tiers, originally designed to be answered by writing Google Earth Engine code whose output is compared against deterministic reference answers; it tests quantitative Earth-observation analysis. UnivEARTH \citep{kao2026univearth} comprises 140 yes/no questions derived from NASA Earth Observatory articles, where the agent must discover appropriate datasets from over 400 imagery collections in the Earth Engine catalog and derive an answer; it tests data discovery and end-to-end reasoning. We adapt both by presenting each item as a problem to solve rather than as code to write, which removes the dependence on a particular implementation route and lets systems with different tool sets be compared (Section~\ref{sec:benchmarks}).

\section{Evaluated Systems}\label{sec:systems}

To quantify what GRA contributes and to separate the value of its geospatial capabilities from that of the orchestration built around them, we compare four system configurations powered by the same underlying Gemini model: a standalone LLM that relies solely on its built-in knowledge; the same LLM augmented with basic tools (Grounding with Google Search, Grounding with Google Maps, and Code Execution); the same LLM given GRA's complete skill and tool suite but none of its orchestration; and GRA itself, which adds upfront planning and execution guidance. Because the model is held fixed across all four, differences in performance can be attributed to capabilities and orchestration rather than to the model. We repeat the full comparison with Gemini 3.1 Pro and Gemini 3.8 Flash to examine how model choice affects both performance and cost.

\textbf{Terminology.} Throughout this paper, we use the following terms consistently: a \emph{query} is the natural-language input from the user; an \emph{orchestration harness} is the software layer that manages tool access, memory, and the agent's multi-step execution loop; an \emph{agent} is an LLM together with its orchestration harness, which lets it answer a query through multiple steps of tool use; a \emph{trajectory} is the sequence of actions (tool calls, code execution, intermediate reasoning) taken by the agent; \emph{upfront planning} is the formulation of an explicit plan before any geospatial tool is invoked, which then guides a step-by-step execution loop that may deviate from the plan as tool outputs arrive; a \emph{reactive} (ReAct-style; \citealp{yao2023react}) agent, by contrast, starts without a plan and chooses each action only after observing the result of the previous one; \emph{execution guidance} is the set of agent-level instructions that governs the execution stage: how the agent adapts its plan as tool outputs arrive and how it handles imperfect information, for example when to stop retrying a data source or to substitute a proxy (Section~\ref{sec:eaia}), as distinct from the per-skill usage guidance that is part of the skills themselves; a \emph{golden workflow} is an expert-defined reference plan specifying expected data sources, analysis steps, and output characteristics (the autorater prompt in Appendix~\ref{app:autoraters} refers to this as the \emph{golden plan}); and an \emph{autorater} (or LLM-as-judge) is an LLM-based scoring system that evaluates agent outputs, whether by comparing final answers against ground truth references or by assessing full trajectories against golden workflows.

\subsection{The Geospatial Reasoning Agent}\label{sec:eaia}

\begin{figure}[t]
\centering
\includegraphics[width=\textwidth, trim=19 52 49 40, clip]{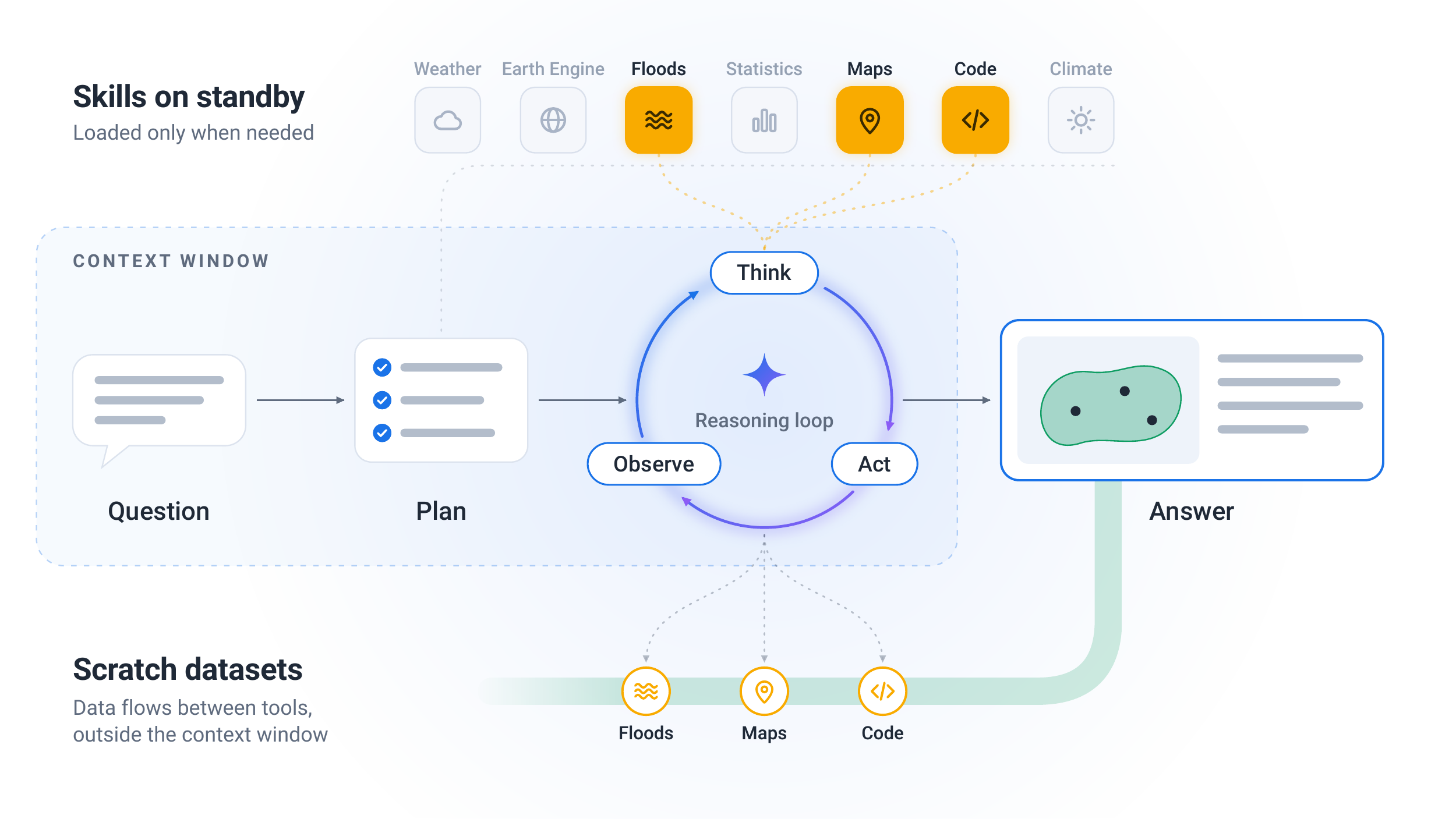}
\caption{Overview of GRA. Skills stay on standby and are loaded only when a query needs them. The agent first sets out a plan, then carries it out in a reasoning loop, while intermediate datasets pass between tools outside the model's context window.}
\label{fig:eaia-arch}
\end{figure}

GRA is a Gemini-powered agent that brings a wide range of geospatial capabilities together behind a single natural-language interface. It follows a domain-oriented architecture, organizing those capabilities as specialized tools, each covering one domain: Google Earth Engine \citep{gorelick2017google} for satellite and geospatial data access; Data Commons \citep{guha2023datacommons, datacommons2025} for socioeconomic and demographic statistics; Google Maps Data for points of interest and other geographic data; Who's On First \citep{whosonfirst} and other open datasets for administrative boundaries; WeatherNext ensemble forecasting \citep{price2025gencast, google2025weather}, Google's experimental cyclone model (FNV3, via Weather Lab) \citep{alet2025cyclone}, and operational and urban flash flood forecasting \citep{nearing2024floods, nevo2022floods} for environmental prediction; the Remote Sensing Foundations models \citep{barzilai2025rsfm} for quick, country-scale, open-vocabulary search and classification over Google's overhead imagery; a domain skill for long-term climate risk; TimesFM \citep{das2024timesfm} for time-series forecasting; and on-the-fly spatial prediction powered by the Population Dynamics Foundation Model \citep{agarwal2024pdfm} and AlphaEarth Foundations \citep{brown2025alphaearth}, which encode demographic, socioeconomic and surface patterns into reusable representations that can be quickly adapted to new prediction tasks. Table~\ref{tab:eaia-pillars} summarizes the capability suite across five domain pillars (indicating assets developed at Google with *), and Appendix~\ref{app:tools} catalogs the individual capabilities. In the evaluated runs, GRA loaded 23 skills and called 62 distinct domain and workspace tools. Figure~\ref{fig:eaia-arch} gives an overview of GRA's components and flow: upfront planning, on-demand skill loading, analysis of data held outside the model's context, and a final response grounded in the results.

\begin{table}[htbp]
\centering
\footnotesize
\setlength{\tabcolsep}{4pt}
\caption{Overview of the GRA capability ecosystem across five domain pillars (* denotes models or datasets developed at Google).}
\label{tab:eaia-pillars}
\begin{tabularx}{\textwidth}{>{\raggedright\arraybackslash}p{2.75cm}>{\raggedright\arraybackslash}X>{\raggedright\arraybackslash}X}
\toprule
\textbf{Capability Pillar} & \textbf{Specialized Models \& Data Assets} & \textbf{Representative Analytical Operations} \\
\midrule
\textbf{Imagery \& Earth Observation} & AlphaEarth Foundations* (annual satellite embeddings); Copernicus DEM GLO-30; Earth Engine* catalog imagery (e.g., Sentinel-1/2, Landsat, MODIS); Remote Sensing Foundations models* & Embedding-based surface change detection; terrain slope; country-scale open-vocabulary search \& classification over overhead imagery \\
\rowrule
\textbf{Weather, Climate \& Hazards} & WeatherNext* (ensemble forecast statistics); Google experimental cyclone model*; Google Flood Forecasting* \& Google Runoff Reanalysis \& Reforecast (GRRR)*; Google Maps Data*; ERA5/ERA5-Land, NASA GPM IMERG, PRISM; NOAA IBTrACS; NEX-GDDP-CMIP6, WRI Aqueduct, JRC GloFAS, USDA Wildfire Risk to Communities; GFPLAIN250m, JRC Global Surface Water & Point weather \& cyclone track/wind forecasting; flood forecasts, inundation \& gauge return periods; floodplain screening; historical weather statistics \& anomalies; climate-hazard projections to 2100 \\
\rowrule
\textbf{Population, Health \& Socioeconomics} & Data Commons* (e.g., US Census ACS, CDC, FEMA NRI, World Bank); PDFM* (Population Dynamics embeddings) & Demographic, economic \& health statistics across administrative levels; regional similarity search \\
\rowrule
\textbf{POIs, Boundaries \& Mobility} & Google Maps Data*; Who's On First, US Census TIGER/Line \& ZCTA & Boundary resolution (countries to census tracts) \& spatial filtering; POI search \& counting; routes, travel-time matrices \& isochrones; rooftop solar potential \\
\rowrule
\textbf{Prediction \& Spatial Compute} & Scratch datasets with sandboxed Python; Google Earth Engine* (general-purpose analysis of any catalog dataset); TimesFM*; PDFM*; Planetary Prediction Engine \citep{ma2026}, which combines them with AlphaEarth Foundations* embeddings and Data Commons* data & Automated data and model selection \citep{ma2026}; out-of-context data handling \& custom Python analysis, including custom Earth Engine scripts; CRS-aware buffering, distances \& spatial joins; regional statistics over gridded datasets; time-series forecasting; regional prediction \& super-resolution \\
\bottomrule
\end{tabularx}
\end{table}

Two changes since the original system \citep{earthai2025} shape how the agent behaves. The first is a shift to a skills-based architecture built around progressive disclosure (Appendix~\ref{app:progd}). In the earlier system every tool the agent could use had to be declared in every request to the model. Even before this approaches the model's context limit, a large set of tool declarations becomes hard for the model to navigate, so each added domain makes every other domain marginally harder to use, consistent with reports that tool-selection accuracy drops as the number of declared tools grows \citep{gan2025ragmcp, kate2025longfunceval}. In contrast, capabilities are now packaged as skills, each pairing a set of domain tools with written guidance on how to use them. The agent sees only a listing of the skill frontmatter (name and description) until it decides which to load. The effect is that the capability surface can grow well beyond what the model could reliably handle if every tool were declared at once, domains can be added or withdrawn independently, and the agent spends its context on the few capabilities a given query actually needs. In addition, large intermediate data is handled on the same principle: satellite rasters, geographic boundaries and tabular results are held as scratch datasets in an external workspace, improving on the design in \citet{earthai2025}, and passed between tools by reference, with the agent inspecting them through a dedicated tool and performing custom processing in a Python environment rather than drawing them into context.

The second change concerns how these capabilities are put to use: following planning-first agent designs \citep{xu2023rewoo, kim2024llmcompiler}, GRA now plans before it acts. The previous agent chose its next action after seeing the result of the current one, a reactive pattern that works well when the required sequence of operations is evident from the question, but that otherwise lets an early, possibly arbitrary tool choice shape the rest of the trajectory \citep{yu2026webanchor}. Instead, the new implementation begins every query with a dedicated planning stage that reasons over available capabilities to assess what it can and cannot do, and sets out the stages of retrieval and processing needed to reach an answer; geospatial tools are used only in the execution stage that follows. The granularity of that plan scales with the problem: a straightforward single-domain question resolves to a single milestone, while an open-ended analytical query is decomposed into a multi-stage workflow that coordinates retrievals across domains and the spatial joins that follow. Execution then treats the plan as an adaptive roadmap rather than a script, deviating from it when tool outputs reveal a more direct route or a gap in data coverage. Alongside the plan, a set of behavioral instructions governs how the agent handles imperfect information: it is directed not to fall back to web search when a dedicated data source is available, to treat an authoritative empty result as a finding rather than a failure, to treat a data request as a dead end after more than three unsuccessful calls to the same tool, and, when operating in the best-effort mode used throughout our evaluation (Section~\ref{sec:setup}), to substitute the closest defensible proxy where an ideal measurement is unavailable and to document any proxy substitutions, spatial/temporal bounds, and uncertainty in its synthesized response. We refer to this adaptive execution of the plan and these behavioral instructions together as \emph{execution guidance}.

\subsection{Evaluated Configurations}

\begin{figure}[t]
\centering
\includegraphics[width=\textwidth, trim=30 64 31 76, clip]{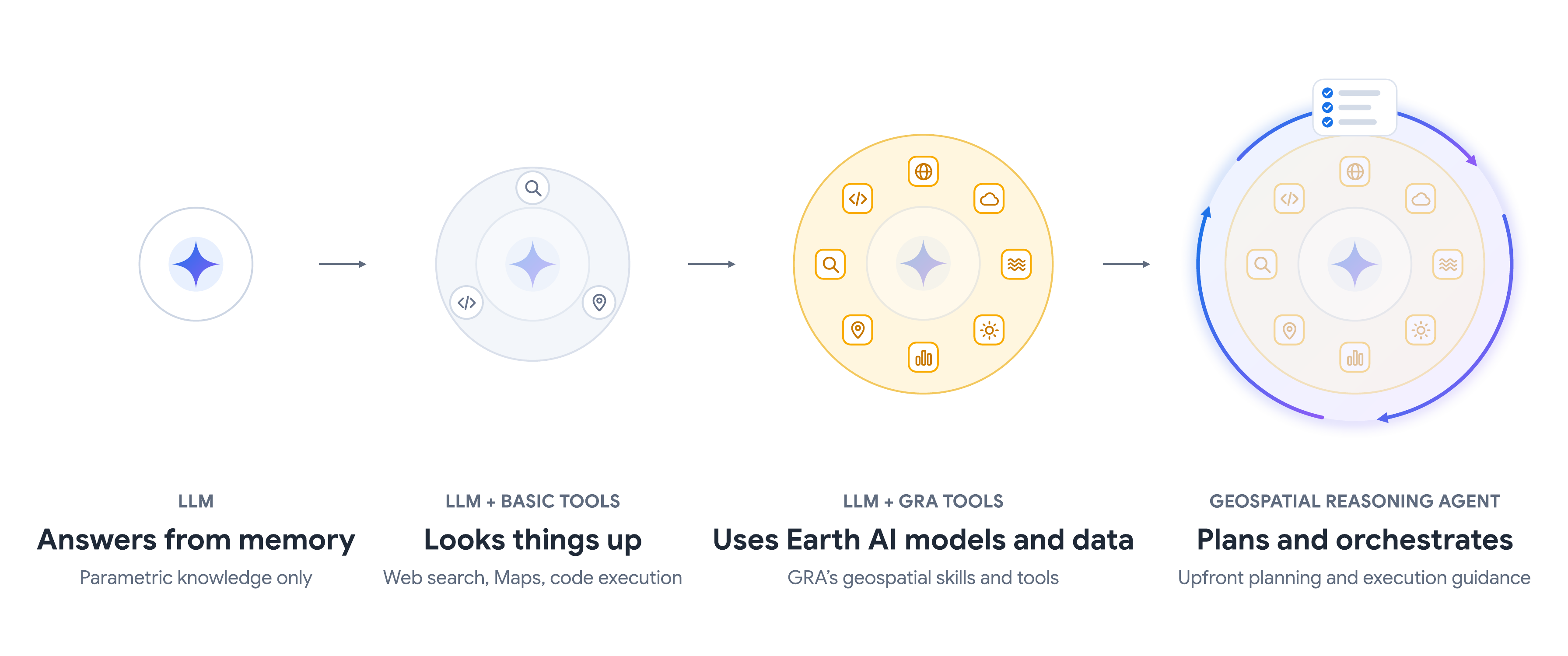}
\caption{The four evaluated configurations. Each adds one layer to the same Gemini model: basic tools, then GRA's geospatial skills and tools, then GRA's upfront planning and execution guidance.}
\label{fig:configs}
\end{figure}

The four configurations below form a progression, each adding one layer to the one before it: general-purpose tool access, then GRA's domain capabilities, then the orchestration built around them (Figure~\ref{fig:configs}). All four run on the same underlying Gemini model.

\textbf{Standalone LLM.} In this configuration, the Gemini model receives the user's query and must answer using only its parametric knowledge. No external tools, data retrieval, or code execution are available. This baseline measures what the model can achieve from parametric knowledge alone.

\textbf{LLM + Basic Tools.} Here, the same Gemini model is augmented with three basic, general-purpose tools: Grounding with Google Search\footnote{\url{https://docs.cloud.google.com/gemini-enterprise-agent-platform/models/grounding/grounding-with-google-search}} for web retrieval, Grounding with Google Maps for place and location lookups, and a sandboxed Python execution environment for on-the-fly computation. The Python sandbox has no network access\footnote{Gemini's built-in Code Execution tool: \url{https://docs.cloud.google.com/gemini-enterprise-agent-platform/models/tools/code-execution}.}; the agent is instructed to route all information retrieval through Search and Maps and to use Python strictly for offline mathematical and tabular analysis. Models do not always follow this instruction, and the resulting failure mode is discussed in Section~\ref{sec:setup}. We deliberately evaluate this configuration rather than provisioning an internet-connected Python environment or pre-installing Earth Engine libraries, for two reasons. First, network access would blur the boundary this study is designed to measure. Our question is what domain-specific geospatial capabilities and orchestration add over general-purpose tooling, and a baseline that can fetch arbitrary remote resources can also fetch geospatial data, whether through an authenticated Earth Engine SDK or by reading cloud-optimized rasters directly from public buckets. Second, allowing ad-hoc requests to public REST APIs (e.g., Overpass, Nominatim, Census) would make evaluation scores reflect third-party rate limits, IP blocks, and socket timeouts rather than model reasoning. This configuration is also consistent with widely deployed LLM code interpreters, which are network-isolated by default.

\textbf{LLM + GRA Tools.} This configuration gives the same Gemini model GRA's skill and tool suite, as described in Section~\ref{sec:eaia}, within a standard reactive loop. The model sees the same skill catalog as GRA, loads skills on demand, and uses the same scratch datasets, but receives only minimal instructions and no planning stage, selecting each action after observing the result of the previous one. It thus differs from GRA only in the orchestration layer: the planning stage and the execution guidance described in Section~\ref{sec:eaia}. Comparing the two measures the contribution of this layer as a whole.

\textbf{Geospatial Reasoning agent (GRA).} The complete system as described in Section~\ref{sec:eaia}: the same capabilities as the previous configuration, together with upfront planning and the execution guidance built around it.

\subsection{Models}\label{sec:models}

We evaluate all four configurations using two Gemini model versions: Gemini 3.1 Pro, a high-capability model designed for complex multi-step reasoning, and Gemini 3.8 Flash, a more recent model from a later generation, optimized for lower latency and cost. Note that these two differ in capacity tier, in release generation, and in their default thinking level (Section~\ref{sec:setup}), so differences between them reflect the combined effect of these factors rather than the cost of choosing a lighter model. Comparing the same system configurations across both models lets us assess whether the patterns we observe are robust to model choice, and what the practical tradeoff looks like for a product deployment decision today.

\subsection{Experimental Setup}\label{sec:setup}

\textbf{Query submission.} All configurations and models receive identical queries. We use default decoding settings, including default temperature, and each model's default thinking level: high for Gemini 3.1 Pro and medium for Gemini 3.8 Flash.

\textbf{Best-effort mode.} Every query, in every configuration, is prefixed with ``Do the following in `best effort' mode:'', so that refusals are not scored as reasoning failures. By default, GRA is instructed to abstain when a required tool or dataset is unavailable. This is appropriate in deployment where ``failing fast'' is desirable to avoid expensive operations if a dead end seems likely, but on open-ended benchmarks, whose queries may reference specialized, proprietary, or regionally unavailable data, it would confuse gaps in tool coverage with gaps in analytical capability. In GRA, the prefix activates instructions to plan around the closest viable substitute, such as an Earth Engine collection, a Data Commons table, or published literature found on the web, and, if a step still fails, to complete the rest of the analysis and report the substitutions made and the resulting uncertainty. The other configurations receive the same prefix but do not have GRA instructions, so for them this prefix acts only as an instruction not to decline. The prefix is not shown to the autoraters, which evaluate each response against the original query.

\textbf{Repeats and variance.} Every query is run three times under each configuration and model. Each repeat produces a complete set of per-query scores, which we aggregate into a single benchmark score, yielding three benchmark scores per configuration. We report the mean and standard deviation across these three. The reported standard deviation therefore reflects run-to-run variability of the aggregate score, not variance across individual queries.

\textbf{Execution limits.} Every run is subject to two limits: a maximum of 100 LLM calls and a wall-clock limit of 30 minutes. Runs exceeding either limit, or the model's 1M-token context window, are aborted and receive a score of zero. Runs that end in an agent-side failure are likewise scored zero, and all such runs remain in the denominator, with their cost included in the reported cost figures. These two limits apply uniformly across all configurations.

\textbf{Baseline harness safeguards.} In our development runs, the model in the LLM + Basic Tools configuration tended to assume that some combination of its tools could answer any query, and rarely concluded otherwise. Without explicit guidance, it often wrote Python that calls external APIs or initializes Earth Engine. In our network-isolated sandbox these calls always fail, and the model spends its remaining budget retrying. Similarly, when the required data is not published on the open web, the model keeps reformulating searches instead of concluding that the data is unavailable. In both cases, the run exhausts its call or time budget and is aborted with a score of zero, which reflects harness friction rather than analytical capability. We therefore add two controls to the LLM + Basic Tools configuration. First, both the orchestrator and the code execution component are told that the Python environment has no network access and no Earth Engine libraries, so retrieval goes through Search and Maps and Python is used only for offline computation. Second, Search is capped at 25 calls per query, with the remaining quota reported in each tool response; this stops the search loop well before the 100-call or 30-minute limit, so the model answers from what it has retrieved instead of being aborted with a zero score.

\section{Benchmarks}\label{sec:benchmarks}

We evaluate the systems described in Section~\ref{sec:systems} on four benchmarks that together span expert-level geospatial reasoning, controlled multi-tool orchestration, and community-developed evaluation tasks.

Two are developed internally and play complementary roles. GeoExpertQueries tests systems on the kind of complex, open-ended problems that practitioners bring to geospatial analysis, which often admit no single correct answer. GeoKTools complements it: its queries are deliberately synthetic and generated from expert-written templates with deterministic answers, so that final results can be checked directly and the number of required tool types can be varied by design. To enable others to build on this work, we publicly release the development splits of both internal benchmarks in full, including the consensus golden workflows and ground-truth answers (Section~\ref{sec:data-release}).

The remaining two benchmarks, CBGB and UnivEARTH, were developed independently of our system and provide an external check on benchmarks we designed ourselves. As we show in Section~\ref{sec:results}, UnivEARTH proves to be saturated, with even a standalone LLM scoring 87 to 94. This illustrates how quickly a public benchmark can lose its discriminative power once released, a general property of public benchmarks in this domain. Table~\ref{tab:benchmarks} summarizes the entire suite.

In addition to task scores, we report the estimated inference cost per query for each system configuration, computed from input, cached-input, output and thinking token counts, each at the model pricing rate current at evaluation time (October 2026; for Gemini 3.1 Pro this includes the higher rate charged for requests above 200K tokens), and including the cost of failed runs (Section~\ref{sec:setup}). The figures cover model inference only, not the cost of external data and compute services such as Earth Engine. This enables cost-accuracy tradeoff analysis across configurations and models (Section~\ref{sec:cost}).

\begin{table}[htbp]
\centering
\small
\setlength{\tabcolsep}{4pt}
\caption{Summary of the benchmark suite.}
\label{tab:benchmarks}
\begin{tabularx}{\textwidth}{l l C{1.7cm} >{\raggedright\arraybackslash}X >{\raggedright\arraybackslash}X}
\toprule
\textbf{Benchmark} & \textbf{Type} & \textbf{Queries (dev/test)} & \textbf{Evaluation} & \textbf{Primary Focus} \\
\midrule
\textbf{GeoExpertQueries} & Internal, open-ended & 25 / 76 & Golden workflows + tool-agnostic autorater & Real-world geospatial reasoning \\
\rowrule
\textbf{GeoKTools} & Internal, controlled & 85 / 220 & Ground-truth answers + autorater & Multi-tool orchestration \\
\rowrule
\textbf{CBGB} & External & --- / 44 & Deterministic answer match (original criterion) & Quantitative Earth-observation analysis \\
\rowrule
\textbf{UnivEARTH} & External & --- / 140 & Ternary answer match & Data discovery and end-to-end reasoning \\
\bottomrule
\end{tabularx}
\end{table}

\subsection{GeoExpertQueries}\label{sec:GeoExpertQueries}

To evaluate performance on realistic, complex geospatial problems of the kind posed by domain practitioners, we curated a benchmark of queries grounded in real-world use cases and refined by domain experts, spanning diverse geospatial domains. These queries are deliberately open-ended and sometimes underspecified, mirroring the way geospatial problems arise in practice: they require resolving ambiguities, identifying and integrating data from multiple sources, and synthesizing multi-step analyses, often with no single correct answer.

\subsubsection{Construction}

The queries were derived from real-world use cases encountered in engagements with practitioners: initial drafts were generated with Gemini 3.0 Flash from notes and documents describing these use cases, and were then reviewed and edited by Earth AI team members with geospatial domain expertise to ensure that each represents a complex, practice-inspired geospatial analysis task. For each query, three independent human experts\footnote{BSc graduates in geospatial fields with a few years of industry experience, MA candidates in geospatial fields, and holders of MSc or PhD degrees in geospatial fields. Experts with the same profile rated the traces in the autorater validation (Section~\ref{sec:ar-validation}).} created golden workflows describing the expected sequence of reasoning steps, data sources, analysis operations, and output characteristics. Importantly, these experts worked independently of one another and blind to GRA and its tool suite, ensuring that the golden workflows reflect domain best practices rather than any particular platform's capabilities.

The three workflows per query were then merged automatically, using an LLM-based procedure, into a single consensus golden workflow that captures the core requirements while accommodating legitimate methodological variation. For four queries the three expert workflows proposed methodologically incompatible approaches and could not be reconciled into a single reference. For example, for one query on water quality, two experts relied on satellite imagery to measure turbidity, while the third expected only local measurements taken at the water intake locations. These queries retain multiple alternative golden workflows, and the autorater scores a response against the closest one, as described below. We regard these cases as informative rather than problematic: they are direct evidence that complex geospatial problems can admit more than one defensible analytical approach, which is one of the reasons evaluation in this domain is difficult. Of the 132 queries originally authored, those determined to be clearly unsolvable, for example requiring data that is unavailable or proprietary, were filtered out, leaving 101 queries.

\subsubsection{Domain coverage and dev/test split}

The 101 queries span four geospatial domains. Queries were split at random within each domain into dev (25) and test (76) sets:

\begin{itemize}
  \item \textbf{Natural Hazards \& Crisis Resilience} (11 dev / 35 test): flood risk, wildfire assessment, natural disaster preparedness and response, and related climate and environmental monitoring.
  \item \textbf{Public Health} (5 dev / 17 test): disease risk mapping, health indicator analysis, environmental health determinants, and demographic vulnerability.
  \item \textbf{Site Selection \& Urban Planning} (8 dev / 20 test): commercial siting, infrastructure planning, land-use suitability, and real estate analytics.
  \item \textbf{Operational \& Commercial Utilities} (1 dev / 4 test): a residual category of cross-domain queries spanning logistics, resource management, and other applied use cases.
\end{itemize}

The fourth category is small, with only four test queries, so per-domain results for it should be read as indicative only. The development set is used to calibrate the evaluation infrastructure and to refine prompts, including the autorater prompt. Table~\ref{tab:geoexpert-examples} shows one example query from each of the three largest domains. The reported results are based on the test set.

\begin{table}[htbp]
\centering
\small
\setlength{\tabcolsep}{4pt}
\caption{Example queries from GeoExpertQueries, drawn from the publicly released development split. Appendix~\ref{app:examples} explains each example in plain language.}
\label{tab:geoexpert-examples}
\begin{tabularx}{\textwidth}{>{\raggedright\arraybackslash}p{3.6cm}>{\raggedright\arraybackslash}X}
\toprule
\textbf{Domain} & \textbf{Example query} \\
\midrule
\textbf{Natural Hazards \& Crisis Resilience} & Using SAR Interferometry (InSAR) coherence change detection, estimate the structural damage in the `hard-hit' zones of the 2022 Pakistan floods in the Sindh and Balochistan provinces. Differentiate between `inundated' (water) and `destroyed' (rubble). Overlay with poverty maps to prioritize reconstruction grants. \\
\rowrule
\textbf{Public Health} & Assess the physical accessibility of emergency obstetric care in Sierra Leone by calculating travel times over the Dynamic World land cover friction surface. \\
\rowrule
\textbf{Site Selection \& Urban Planning} & For the city of Buenos Aires, determine the `Building Footprint' area vs. available land area. \\
\bottomrule
\end{tabularx}
\end{table}

\subsubsection{Evaluation}

Performance is assessed using an LLM-as-judge autorater \citep{zheng2023judging}, powered by Gemini 3.8 Flash, that receives the user query, the system's response including its execution trajectory, and the consensus golden workflow. Because these queries are open-ended with no single correct answer, the autorater rates the analysis the system actually executed rather than checking the final answer against a fixed ground truth: it examines the system's tool calls, the data they return, and the processing applied to that data, and judges how much of the analytical value specified by the golden workflow they deliver. For the small number of queries that retain multiple alternative golden workflows, as noted above, the response is scored against the closest matching alternative. Each response is scored by three independent autorater runs, each returning 0, 0.5, or 1, and the score for that response is their average.

Before scoring, traces are compacted to remove execution noise while preserving the sequence of analysis steps and the final response. The same processing is applied to all system configurations, so that no configuration is advantaged or penalized by how its output is presented to the judge.

A key design consideration is cross-configuration fairness. Since we compare systems with very different tool sets, the autorater treats the golden workflow as a reference rather than a checklist: an approach counts as following it if a geospatial expert would accept it as answering the query equally well, without favoring any particular dataset, tool, platform or provider. A step is credited if it is carried out by any means capable of performing it -- whether using a domain-specific geospatial tool, an equivalent dataset, or a custom Python calculation; this is what we mean by tool-agnostic scoring. In particular, LLM + Basic Tools, which can gather information through web search, is not penalized merely for not using domain-specific tools.

The autorater follows a small set of general rules; the full prompt is given in Appendix~\ref{app:autoraters}. First, only executed work counts: a step that the system merely describes or promises earns no credit, and a step does not count if its inputs, such as coordinates or published statistics, are hardcoded from memory rather than retrieved by a tool. Second, deviations from the golden workflow are weighed by their impact on the answer rather than by their number. Deviations include using a proxy for the required data (for example, a whole district's population in place of the population inside the flood extent), skipping a core step, and applying a simpler method that answers a less precise question. A response receives full credit if its deviations cause at most a slight loss of usefulness to a geospatial expert, half credit if they reduce it significantly, and no credit if they are fatal. Third, a response receives no credit if it offers only a plan, ends without a final answer, performs no analysis on data retrieved by tools (for instance, relying on memory or a generic web lookup where geospatial analysis is required), or fabricates tool outputs. Conversely, the autorater does not penalize reasonable choices for parameters the query leaves open, such as the study area or time window; skipped optional steps; technical details such as projections or units; missing maps or charts; or a failed tool call from which the system recovers with an equally good alternative. Whether a given substitution is defensible is ultimately a domain judgment, which these rules aim to approximate. The following subsection reports how well the autorater's overall scores agree with expert ratings; a finer-grained analysis of which specific substitutions experts would accept remains future work.

\subsubsection{Autorater validation}\label{sec:ar-validation}
We sampled 69 traces on test-set queries, 57 produced by GRA and 12 by LLM + GRA Tools (35 with Gemini 3.1 Pro, 34 with Gemini 3.8 Flash), spanning low, middle, and high autorater scores. Three independent domain experts rated each trace on a three-point scale, Good, Partial, or Poor, against the consensus golden workflow, yielding 207 annotations. Annotators did not see the autorater score and did not know which model or configuration produced the trace. For analysis, ratings were mapped to 0, 0.5, and 1 and averaged across the three experts; autorater scores were averaged across three independent passes.

The autorater agrees with the experts about as well as the experts agree with one another. Krippendorff's $\alpha$ (interval) among the three experts is 0.56 (95\% bootstrap CI 0.42 to 0.67); adding the autorater as a fourth rater gives 0.57 (0.45 to 0.67), and replacing any one expert with the autorater gives 0.59, 0.55, or 0.59. Pairwise, the autorater's agreement with each expert ($\alpha$ = 0.55, 0.64, 0.57) is comparable to the agreement between pairs of experts (0.59, 0.53, 0.56). Spearman correlation between the autorater score and the mean expert rating is 0.72 (0.60 to 0.82). On the 0--1 scale, the autorater score and the expert mean differ by 0.20 per trace on average, and these differences go in both directions about equally (mean of autorater minus experts: +0.01), so the autorater is neither systematically harsher nor more lenient than the panel. Disagreements are small. For 87\% of traces the autorater score lies within 0.33 of the expert mean, that is, less than one rating category (0.5) apart, and there are no outright reversals: no trace that the autorater scored above 0.67 received a mean expert score below 0.33, or vice versa.

\subsection{GeoKTools}\label{sec:geoktools}

\begin{figure}[t]
\input{figures/fig_geoktools_body}
\caption{\textbf{End-to-end construction of a triple-tool evaluation query example ($K=3$).} Sampled argument values (e.g., geographic regions, landmarks, years) populate the template (see (1)) to generate concrete query instances (see (3)), and serve as input variables to the reference solver logic (see (2)). The solver then executes a structured multi-tool workflow, combining spatial analysis, POI lookup, and public statistics, to deterministically compute ground-truth answers (see (4)) for automated scoring.}
\label{fig:geoktools}
\end{figure}

To measure how performance varies with the number of tools a task requires, we constructed a benchmark of templated queries with verifiable ground-truth answers, organized by $K$, the number of distinct tool types (e.g., spatial analysis, weather forecasting, demographic lookup) required to answer each query. Across all templates, the benchmark draws on 24 tool types. $K$ thus serves as a potential proxy for task complexity, and this design lets us observe how performance changes as $K$ increases from 1 (a single tool) to 4 (a four-tool chain). Note that $K$ controls the number of tool types rather than the intrinsic difficulty of each step, and templates at a given $K$ differ in other respects. Each query is designed so that the correct answer can be computed deterministically from the query parameters using a reference Python solver, enabling reliable evaluation without the ambiguity of open-ended assessment. Since these queries are constructed around GRA's tool suite, in the sense that the templates were written to require operations those tools support, the advantage of the two configurations equipped with GRA's tools is expected, although neither receives any information about which tools are relevant. The more informative findings concern how far baselines can compensate through parametric knowledge or basic tools.

\subsubsection{Construction}

Domain experts developed 112 query templates (see (1) in Figure~\ref{fig:geoktools}), organized by the number of tool types required. For each template, a reference solver ((2) in Figure~\ref{fig:geoktools}) executes a predefined tool-calling workflow and data-processing pipeline to compute the ground-truth answer for any given set of input parameters. Concrete query instances were then generated by randomly sampling parameter values (e.g., geographic regions, dates, statistical indicators, and thresholds), simultaneously populating the templates and producing verified question-answer pairs ((3) and (4) in Figure~\ref{fig:geoktools}). Some templates were set aside during development for technical reasons, leaving 92 templates (9/64/11/8 for $K=1/2/3/4$), from which we sampled 305 queries.

\subsubsection{Dev/test split}

Since each template generates multiple query instances (by sampling different parameter values), queries derived from the same template share the same underlying structure and solving logic. To prevent data leakage, the dev/test set split was performed at the template level: all instances of a given template are assigned entirely to either the dev or the test set. The development set contains 18 templates (2/12/2/2 for $K=1/2/3/4$), yielding 85 queries (10/50/15/10). The held-out test set contains 74 templates (7/52/9/6 for $K=1/2/3/4$), yielding 220 queries (30/110/50/30). The development set is used to calibrate evaluation infrastructure (e.g., prompt selection, autorater validation, and end-to-end pipeline checks). All reported results are based on the test set.

\subsubsection{Evaluation}

Each query contains one to seven blanks, whose reference values (strings such as a country name, numbers, or both) are computed by the reference solver, and the system is instructed to respond with a JSON list filling in these blanks. A response that exactly matches the reference scores 1.0 without invoking a judge, and an empty response scores 0. Otherwise, an LLM judge (Gemini 3.8 Flash) scores each element between 0 and 1. Rather than applying a fixed tolerance, it assesses each discrepancy in the context of the query's units and domain scale, so that near-misses, such as equivalent place names or small differences due to rounding or data sources, lose little or no credit, while larger errors lose more. If the response identifies the wrong primary entity, elements that depend on it score 0. The query score is the mean of its element scores. The full autorater prompt is provided in Appendix~\ref{app:autoraters}.

\subsection{Data Release}\label{sec:data-release}

To support reproducibility and further research, we publicly release the development splits of both internal benchmarks at \url{https://github.com/google-research/geospatial-reasoning-benchmarks}. For GeoExpertQueries we release the 25 queries with their golden workflows (the consensus workflow, or for one query the alternative expert workflows) and the domain labels. For GeoKTools we release the 85 queries (10/50/15/10 for $K=1/2/3/4$), drawn from 18 templates, with their ground-truth answers. For each template we also give a specification of its reference solver (i.e., a step-by-step description of the computation). We also release the scoring code for both benchmarks. For GeoExpertQueries this comprises the autorater prompt and a runner. For GeoKTools it is an LLM-as-judge runner that follows the protocol of Section~\ref{sec:geoktools}, plus a deterministic scorer that needs no model calls. Code is released under Apache 2.0 and benchmark content under CC BY 4.0. Ground-truth values derived from Google Maps Data or third-party data sources remain subject to those sources' terms.

The autorater prompts are also provided in Appendix~\ref{app:autoraters}. Because ground-truth values are computed from live data sources, the released answers reflect a snapshot taken on Sep 23, 2026; re-running the solvers later may yield different values.

We withhold the test splits deliberately: as Section~\ref{sec:res-univearth} shows, a public benchmark can lose its discriminative power once the source material its items are derived from is widely published or accessible through model training data, and holding back the test items preserves them as an uncontaminated signal. The released template specifications also let others build new instances of the same form.

\subsection{External Benchmarks}\label{sec:external-benchmarks}

To complement our internally developed benchmarks, we include two externally developed benchmarks of geospatial analysis tasks with deterministic answers, both built around Earth Engine data. CBGB \citep{cardille2025cbgb} poses well-specified geospatial questions whose answers are specific numeric values, testing whether a system can carry out a prescribed analysis correctly. UnivEARTH \citep{kao2026univearth} is designed to add an upstream challenge: the system must first identify the appropriate imagery collection from a catalog of over 400 options before it can perform the computation, combining data discovery with analysis. In both cases we present the system with the question itself rather than instructing it to produce Earth Engine code, consistent with how we evaluate our internal benchmarks.

\subsubsection{CBGB}\label{sec:cbgb}

CBGB \citep{cardille2025cbgb} (the Cloud-Based Geospatial Benchmark) comprises 45 expert-curated geospatial questions, each with a single numerical answer. The questions span deforestation monitoring, wildfire damage assessment, urban heat island analysis, crop health evaluation, and other Earth observation topics, curated by remote sensing and GIS practitioners from the widely used \emph{Cloud-Based Remote Sensing with Google Earth Engine} textbook \citep{eefa2024} and from original problems developed by domain experts. Each question is assigned a difficulty tier: Easy (15), Intermediate (20), or Difficult (10).

CBGB was originally designed as an Earth Engine code-generation benchmark, in which systems are instructed to produce Python code that is then executed against the Earth Engine catalog. We run it differently in two respects: systems receive the question as a geospatial problem to solve, with no instruction about which tools or platform to use, matching how we evaluate our internal benchmarks; and each system gets a single attempt per question, whereas the original protocol allowed up to three. Our results are therefore not directly comparable with those published in the original paper. Only GRA and LLM + GRA Tools have native Earth Engine access; the two baselines must answer from parametric knowledge, web search, or general-purpose code execution.

Following the original CBGB evaluation, our primary metric scores an answer as correct if it matches the reference to two significant figures. This criterion is deterministic and requires no model judgment, which is appropriate for questions with a single numeric answer. For comparison, we also scored the answers with the autorater used for GeoKTools (Section~\ref{sec:geoktools}), which awards partial credit that decreases with how far an answer deviates from the reference.

We evaluate on a modified 44-question version of the benchmark rather than the verbatim original release, for two unrelated reasons. First, we exclude one Difficult item. Its reference solution requires roughly ten iterations of a binary search over polygon buffers, each involving a multi-minute Earth Engine reduction, so a single attempt runs for hours and exceeds the execution limits described in Section~\ref{sec:setup}. The underlying reasoning is not unusually difficult, and systems generally reach the correct answer given enough time.

Second, we updated four reference answers for three Intermediate and one Difficult queries. On inspection their original values no longer matched the data: Sentinel-2 imagery had been reprocessed, and in one case the published boundary of a protected area had been redrawn. These are not errors in the original benchmark but consequences of the underlying data changing after it was built.

\subsubsection{UnivEARTH}\label{sec:univearth}

UnivEARTH \citep{kao2026univearth} contains 140 yes/no questions derived from NASA Earth Observatory articles, spanning 13 topics including land, water, atmosphere, and human presence. A typical question asks whether some measurable change occurred, for example whether surface water extent increased in a given region between two years. Answering it properly requires identifying the appropriate imagery collection from over 400 in the Earth Engine catalog, retrieving and processing the relevant data, and reaching a conclusion.

As with CBGB, we present the question directly rather than instructing the system to produce Earth Engine code. Each question is followed by an instruction to return the final answer as a single-element JSON list containing Yes, No, or Inconclusive, and scoring is an exact match of the returned label against the reference answer, with Inconclusive counted as incorrect.

The original UnivEARTH evaluation, using frontier models from early 2025 (e.g., Claude 3.7 Sonnet, o3-mini, and DeepSeek-V3), reported that the best systems reached roughly 33\% accuracy in the grounded setting (with access to Google Earth Engine), below what random guessing would produce. While models achieved 80\% selective accuracy when their Earth Engine scripts succeeded, overall performance was bottlenecked by a more than 58\% execution failure rate driven largely by hallucinated catalog asset names \citep{kao2026univearth}. As we show in Section~\ref{sec:res-univearth}, current models behave very differently, and we therefore report UnivEARTH as a finding about benchmark saturation rather than as a capability measurement.

\section{Results}\label{sec:results}

We report results across all four benchmarks for the four system configurations described in Section~\ref{sec:systems}. Experiments are conducted with Gemini 3.1 Pro and Gemini 3.8 Flash. All results are based on test sets. Each evaluation was run three times, and we report the mean and standard deviation across the three runs. Table~\ref{tab:main-results} summarizes the main results.

\subsection{Main Results}\label{sec:res-main}

\begin{table}[htbp]
\centering
\footnotesize
\setlength{\tabcolsep}{2pt}
\caption{Main results (test set). Scores are 0--100, mean $\pm$ std over three runs. For CBGB, the score is the average over all 44 questions, across its three difficulty tiers. Bold marks the best result in each row and results whose $\pm$ ranges overlap with it.}
\label{tab:main-results}
\begin{tabular}{l *{8}{C{1.5cm}}}
\toprule
\textbf{Benchmark} & \multicolumn{2}{c}{\textbf{LLM}} & \multicolumn{2}{c}{\begin{tabular}[b]{@{}c@{}}\textbf{LLM +} \\ \textbf{Basic Tools}\end{tabular}} & \multicolumn{2}{c}{\begin{tabular}[b]{@{}c@{}}\textbf{LLM +} \\ \textbf{GRA Tools}\end{tabular}} & \multicolumn{2}{c}{\textbf{GRA}} \\
\cmidrule(lr){2-3} \cmidrule(lr){4-5} \cmidrule(lr){6-7} \cmidrule(lr){8-9}
 & \textbf{3.1P} & \textbf{3.8F} & \textbf{3.1P} & \textbf{3.8F} & \textbf{3.1P} & \textbf{3.8F} & \textbf{3.1P} & \textbf{3.8F} \\
\midrule
\textbf{GeoExpertQueries} & 0.0$\pm$0.0 & 0.0$\pm$0.0 & 2.3$\pm$0.8 & 3.3$\pm$0.8 & 50.8$\pm$1.7 & 52.2$\pm$5.4 & 47.6$\pm$2.7 & \textbf{62.7$\pm$2.5} \\
\textbf{GeoKTools} & 37.1$\pm$1.2 & 40.5$\pm$0.7 & 37.8$\pm$1.9 & 40.0$\pm$0.8 & \textbf{84.5$\pm$1.4} & \textbf{86.0$\pm$1.6} & 81.6$\pm$1.4 & \textbf{85.9$\pm$1.5} \\
\textbf{CBGB} & 14.4$\pm$3.5 & 12.9$\pm$1.3 & 13.6$\pm$2.3 & 19.7$\pm$2.6 & 84.8$\pm$1.3 & \textbf{87.9$\pm$1.3} & 84.8$\pm$1.3 & \textbf{86.4$\pm$2.3} \\
\textbf{UnivEARTH} & 86.9$\pm$1.8 & 93.6$\pm$0.7 & \textbf{96.7$\pm$1.8} & \textbf{98.3$\pm$0.8} & \textbf{97.9$\pm$0.7} & \textbf{98.8$\pm$0.4} & 89.5$\pm$0.4 & \textbf{98.1$\pm$0.4} \\
\bottomrule
\end{tabular}
\end{table}

On GeoExpertQueries, GeoKTools, and CBGB, GRA substantially outperforms both baselines. The contrast is starkest on GeoExpertQueries, where the standalone LLM, which has no means of retrieving data, scores 0.0 by construction and LLM + Basic Tools reaches at most 3.3, against 62.7 for GRA with Gemini 3.8 Flash. Evidently, the multi-step analyses these queries demand leave little room for informed guessing. On GeoKTools, the strongest baseline, a standalone LLM with no tools at all, reaches 40.5 against GRA's 85.9, since the templated queries ask for specific factual values that a model can sometimes approximate from parametric knowledge. On CBGB, GRA scores above 80 with both models, while neither baseline exceeds 20.

Comparing LLM + GRA Tools with full GRA separates the contributions of the two layers. GRA's geospatial skills and tools account for most of the gain on all three benchmarks: on their own, they raise GeoExpertQueries from 3.3 to 52.2 with Gemini 3.8 Flash and from 2.3 to 50.8 with Gemini 3.1 Pro; GeoKTools from about 40 to 85--86; and CBGB from below 20 to 85--88. Upfront planning and execution guidance then add 10.5 points on GeoExpertQueries with Gemini 3.8 Flash (52.2 to 62.7), but not with Gemini 3.1 Pro, where GRA scores 3.2 points lower (47.6 versus 50.8), a difference we examine in Section~\ref{sec:res-GeoExpertQueries}. On GeoKTools and CBGB, where each query has a single well-defined answer, the two configurations score within three points of each other, and LLM + GRA Tools is level or slightly ahead (for example, 84.5 versus 81.6 on GeoKTools with Gemini 3.1 Pro, and 87.9 versus 86.4 on CBGB with Gemini 3.8 Flash). A plausible reading is that upfront planning and execution guidance pay off mainly on open-ended problems and matters less when each query has a well-defined answer, although the lack of a gain with Gemini 3.1 Pro shows that the model plays a role as well.

UnivEARTH shows a different pattern. Every configuration scores between 87 and 99, including the standalone LLM, which has no means of accessing Earth Engine data at all. The benchmark no longer distinguishes between systems, as explained in Section~\ref{sec:res-univearth}.

\subsection{GeoExpertQueries: Expert-Level Geospatial Reasoning}\label{sec:res-GeoExpertQueries}

\begin{table}[htbp]
\centering
\footnotesize
\setlength{\tabcolsep}{4pt}
\caption{GeoExpertQueries results by domain (test set). Scores are 0--100, mean $\pm$ std over three runs. The standalone LLM scores 0.0 in every domain with both models and is omitted. Domains: Natural Hazards \& Crisis Resilience; Public Health; Site Selection \& Urban Planning; Operational \& Commercial Utilities. Bold marks the best result in each row and results whose $\pm$ ranges overlap with it.}
\label{tab:geoexpertprompts}
\begin{tabular}{l *{6}{C{1.6cm}}}
\toprule
\textbf{Domain (n)} & \multicolumn{2}{c}{\begin{tabular}[b]{@{}c@{}}\textbf{LLM +} \\ \textbf{Basic Tools}\end{tabular}} & \multicolumn{2}{c}{\begin{tabular}[b]{@{}c@{}}\textbf{LLM +} \\ \textbf{GRA Tools}\end{tabular}} & \multicolumn{2}{c}{\textbf{GRA}} \\
\cmidrule(lr){2-3} \cmidrule(lr){4-5} \cmidrule(lr){6-7}
 & \textbf{3.1P} & \textbf{3.8F} & \textbf{3.1P} & \textbf{3.8F} & \textbf{3.1P} & \textbf{3.8F} \\
\midrule
\textbf{Crisis Resilience (35)} & 1.0$\pm$1.6 & 1.0$\pm$1.6 & 49.4$\pm$1.5 & 45.6$\pm$3.8 & 50.0$\pm$1.4 & \textbf{66.0$\pm$2.6} \\
\textbf{Public Health (17)} & 0.0$\pm$0.0 & 0.0$\pm$0.0 & \textbf{45.4$\pm$8.6} & \textbf{42.5$\pm$10.2} & 39.2$\pm$2.9 & \textbf{55.6$\pm$3.2} \\
\textbf{Site Selection (20)} & 3.6$\pm$0.5 & 6.9$\pm$1.3 & 57.2$\pm$5.3 & \textbf{71.4$\pm$5.0} & 48.1$\pm$6.3 & \textbf{65.6$\pm$3.4} \\
\textbf{Operational (4)} & 18.1$\pm$4.8 & 19.4$\pm$6.4 & \textbf{54.2$\pm$7.2} & \textbf{55.6$\pm$6.4} & \textbf{59.7$\pm$4.8} & 50.0$\pm$0.0 \\
\textbf{All (76)} & 2.3$\pm$0.8 & 3.3$\pm$0.8 & 50.8$\pm$1.7 & 52.2$\pm$5.4 & 47.6$\pm$2.7 & \textbf{62.7$\pm$2.5} \\
\bottomrule
\end{tabular}
\end{table}

GeoExpertQueries presents the most realistic and challenging evaluation in our suite: open-ended queries derived from practitioner use cases and refined by domain experts, with no single correct answer. The complex multi-step analyses it requires leave little room for educated guesses drawn from parametric knowledge or web search. The results reflect this. The standalone LLM scores exactly 0.0 in every domain, under both models: the autorater credits only findings produced by data retrieval or computation, so answers from memory receive no credit. LLM + Basic Tools reaches 2.3 with Gemini 3.1 Pro and 3.3 with Gemini 3.8 Flash. This credit comes from queries in which some steps, such as looking up a flood zone designation or computing a break-even point, can be approximated with a web search or a short calculation; on more than 80\% of the queries it scores 0 in every run. LLM + GRA Tools lifts scores to 50.8 and 52.2, and full GRA reaches 47.6 and 62.7, respectively.

These results suggest that GRA already has many of the building blocks needed for end-to-end geospatial analyses of the kind that domain practitioners perform: identifying and integrating relevant data sources, applying suitable analytical methods, and synthesizing results into coherent answers, covering a substantial share of the steps in the expert-authored golden workflows. That said, absolute scores on this benchmark should be read with some caution, since they depend on the rating methodology: a stricter autorater would likely assign lower scores across the board. Because all configurations are scored by the same autorater, however, we expect the rankings of the configurations, including GRA's lead over LLM + GRA Tools with Gemini 3.8 Flash, to hold.

As Table~\ref{tab:geoexpertprompts} shows, the gain from upfront planning and execution guidance with Gemini 3.8 Flash comes mainly from Natural Hazards \& Crisis Resilience, the largest domain with 35 of the 76 queries (+20.4 points, from 45.6 to 66.0), and Public Health (+13.1 points, from 42.5 to 55.6, albeit with a standard deviation of 10.2 for LLM + GRA Tools), while differences on Site Selection \& Urban Planning (71.4 versus 65.6) and Operational \& Commercial Utilities (55.6 versus 50.0, with only four test items) are within run-to-run variation. With Gemini 3.1 Pro, GRA does not improve on LLM + GRA Tools in any of the three main domains, and scores lower on Site Selection \& Urban Planning (48.1 versus 57.2). Investigation of the traces suggests an explanation: many Natural Hazards \& Crisis Resilience queries ask about specific hazards, environmental conditions, or named events (such as wildfire damage, flood or earthquake impacts, or reservoir and water-quality conditions), and without a plan, LLM + GRA Tools with Gemini 3.8 Flash often treats them as look-up questions. In a third of its runs in this domain, it starts by loading only the web search skill, and most of these runs never turn to geospatial analysis, scoring below 10 on average. GRA, whose plan and instructions steer it toward dedicated data sources rather than web search, largely avoids this pattern. This is the effect described in Section~\ref{sec:eaia}, where an early tool choice shapes the rest of the trajectory.

One plausible explanation for the lack of a gain with Gemini 3.1 Pro, consistent with the traces, is how closely each model follows its plan. An upfront plan is formed from high-level capability descriptions, before any data has been seen, and cannot anticipate every step or tool a query will end up needing. Gemini 3.8 Flash departs from the plan more readily when execution calls for it, for example by writing custom code or calling a tool again, whereas Gemini 3.1 Pro adheres more strictly both to the plan and to GRA's operational rules, such as limiting repeated calls to the same tool and preferring dedicated tools over custom code. In our setting, this flexibility tends to bring Flash's analyses closer to the golden workflows, which often include best-practice steps that the query does not spell out, such as masking water bodies before computing developable land area. GRA's operational rules reflect experience from real-world use, where they help keep the agent efficient and focused, for example by cutting off unproductive repeated calls; in the present evaluation, however, strict adherence to them may keep Gemini 3.1 Pro from carrying out steps that the expert workflows expect. The same trade-off is visible with Gemini 3.8 Flash on Site Selection \& Urban Planning, where the two configurations score within run-to-run variation of each other. There, both configurations run into the same tool failures, such as empty points-of-interest results or an unavailable air-quality history, but LLM + GRA Tools more often recovers by writing custom code or consulting an auxiliary source, whereas GRA tends to complete its planned steps and stop at a weaker proxy. With Gemini 3.8 Flash, the gain from upfront planning and execution guidance is therefore uneven across domains: large where the plan keeps the agent from treating a query as a web look-up, and absent where both configurations run into the same tool failures.

\subsection{GeoKTools: Effect of Tool Complexity}\label{sec:res-geoktools}

Table~\ref{tab:geoktools} breaks down GeoKTools results by the number of required tool types ($K$). Since GeoKTools queries are designed around GRA's geospatial tools, the strong advantage of the two configurations equipped with them is expected. More informative is how each configuration handles increasing complexity, and how much the baselines can recover from parametric knowledge or web search.

\begin{table}[htbp]
\centering
\footnotesize
\setlength{\tabcolsep}{2pt}
\caption{GeoKTools results by number of required tool types, $K$ (test set; query counts in parentheses). Scores are 0--100, mean $\pm$ std over three runs. Bold marks the best result in each row and results whose $\pm$ ranges overlap with it.}
\label{tab:geoktools}
\begin{tabular}{l *{8}{C{1.5cm}}}
\toprule
\textbf{\boldmath $K$} & \multicolumn{2}{c}{\textbf{LLM}} & \multicolumn{2}{c}{\begin{tabular}[b]{@{}c@{}}\textbf{LLM +} \\ \textbf{Basic Tools}\end{tabular}} & \multicolumn{2}{c}{\begin{tabular}[b]{@{}c@{}}\textbf{LLM +} \\ \textbf{GRA Tools}\end{tabular}} & \multicolumn{2}{c}{\textbf{GRA}} \\
\cmidrule(lr){2-3} \cmidrule(lr){4-5} \cmidrule(lr){6-7} \cmidrule(lr){8-9}
 & \textbf{3.1P} & \textbf{3.8F} & \textbf{3.1P} & \textbf{3.8F} & \textbf{3.1P} & \textbf{3.8F} & \textbf{3.1P} & \textbf{3.8F} \\
\midrule
\textbf{\boldmath $K=1$ (30)} & 46.5$\pm$2.5 & 43.7$\pm$3.2 & 47.7$\pm$5.0 & 51.6$\pm$3.4 & 85.0$\pm$2.1 & \textbf{90.4$\pm$3.8} & \textbf{86.2$\pm$7.1} & \textbf{91.2$\pm$2.1} \\
\textbf{\boldmath $K=2$ (110)} & 41.0$\pm$1.7 & 44.0$\pm$0.1 & 39.1$\pm$2.8 & 41.8$\pm$1.5 & \textbf{85.1$\pm$1.9} & \textbf{85.3$\pm$2.2} & 79.4$\pm$1.6 & \textbf{85.3$\pm$1.1} \\
\textbf{\boldmath $K=3$ (50)} & 28.9$\pm$2.0 & 33.9$\pm$1.3 & 35.8$\pm$2.6 & 35.3$\pm$1.2 & \textbf{83.3$\pm$4.1} & \textbf{86.2$\pm$2.2} & 80.1$\pm$3.6 & \textbf{82.0$\pm$2.6} \\
\textbf{\boldmath $K=4$ (30)} & 27.3$\pm$3.8 & 35.6$\pm$3.6 & 26.9$\pm$2.9 & 29.4$\pm$1.4 & 83.6$\pm$1.1 & \textbf{84.0$\pm$3.3} & \textbf{87.8$\pm$1.6} & \textbf{89.2$\pm$2.3} \\
\textbf{All (220)} & 37.1$\pm$1.2 & 40.5$\pm$0.7 & 37.8$\pm$1.9 & 40.0$\pm$0.8 & \textbf{84.5$\pm$1.4} & \textbf{86.0$\pm$1.6} & 81.6$\pm$1.4 & \textbf{85.9$\pm$1.5} \\
\bottomrule
\end{tabular}
\end{table}

At first glance one might expect baselines to score near zero on these queries, since answering relatively specific geospatial questions, for example about the median age in 2019 in a particular county in Nebraska, would seem to require actual tool-based computation. In practice, however, LLMs possess substantial geographic and demographic knowledge from pretraining. They often follow a reasoning trajectory similar to GRA's, identifying the right types of operations and producing approximate answers drawn from memory or from simple web search rather than from live data retrieval. Because our autorater assigns continuous scores between 0 and 1, crediting near-misses proportionally (Section~\ref{sec:geoktools}), these educated guesses yield partial scores. Adding basic tools does not help: LLM + Basic Tools scores about the same as the standalone LLM (37.8 versus 37.1 with Gemini 3.1 Pro, 40.0 versus 40.5 with Gemini 3.8 Flash).

Baseline scores decline as more tool types are required, from 39--52 at $K=1$ and $K=2$ to 27--36 at $K=3$ and $K=4$, consistent with the compounding difficulty of guessing across several data sources at once. The two configurations with GRA's tools remain far stronger at every level of $K$, and their scores vary little with $K$, staying between 79.4 and 91.2 at every level. As discussed in Section~\ref{sec:geoktools}, $K$ counts tool types rather than intrinsic difficulty, and since the $K=1$, 3 and 4 cells contain only 6--9 templates each, a few templates can shift a cell's score, so differences across $K$ should not be attributed to $K$ alone. Overall, the number of required tool types appears to be a useful but imperfect proxy for task complexity: it tracks difficulty for the baselines, but not for the tool-equipped configurations, which handle four-tool chains about as well as single-tool queries. As noted in Section~\ref{sec:res-main}, upfront planning and execution guidance add no consistent gain on this benchmark.

\subsection{CBGB: External Corroboration}\label{sec:res-cbgb}

\begin{table}[htbp]
\centering
\footnotesize
\setlength{\tabcolsep}{2pt}
\caption{CBGB results by difficulty tier (test set; question counts in parentheses). Scores are 0--100, mean $\pm$ std over three runs. Bold marks the best result in each row and results whose $\pm$ ranges overlap with it.}
\label{tab:cbgb-tiers}
\begin{tabular}{l *{8}{C{1.5cm}}}
\toprule
\textbf{Tier (n)} & \multicolumn{2}{c}{\textbf{LLM}} & \multicolumn{2}{c}{\begin{tabular}[b]{@{}c@{}}\textbf{LLM +} \\ \textbf{Basic Tools}\end{tabular}} & \multicolumn{2}{c}{\begin{tabular}[b]{@{}c@{}}\textbf{LLM +} \\ \textbf{GRA Tools}\end{tabular}} & \multicolumn{2}{c}{\textbf{GRA}} \\
\cmidrule(lr){2-3} \cmidrule(lr){4-5} \cmidrule(lr){6-7} \cmidrule(lr){8-9}
 & \textbf{3.1P} & \textbf{3.8F} & \textbf{3.1P} & \textbf{3.8F} & \textbf{3.1P} & \textbf{3.8F} & \textbf{3.1P} & \textbf{3.8F} \\
\midrule
\textbf{Easy (15)} & 24.4$\pm$3.8 & 22.2$\pm$3.8 & 22.2$\pm$3.8 & 26.7$\pm$0.0 & \textbf{100.0$\pm$0.0} & \textbf{100.0$\pm$0.0} & 95.6$\pm$3.8 & \textbf{97.8$\pm$3.8} \\
\textbf{Intermediate (20)} & 8.3$\pm$5.8 & 11.7$\pm$2.9 & 13.3$\pm$5.8 & 21.7$\pm$2.9 & \textbf{78.3$\pm$5.8} & \textbf{81.7$\pm$2.9} & \textbf{80.0$\pm$0.0} & \textbf{81.7$\pm$2.9} \\
\textbf{Difficult (9)} & 11.1$\pm$0.0 & 0.0$\pm$0.0 & 0.0$\pm$0.0 & 3.7$\pm$6.4 & \textbf{74.1$\pm$6.4} & \textbf{81.5$\pm$6.4} & \textbf{77.8$\pm$0.0} & \textbf{77.8$\pm$0.0} \\
\textbf{All (44)} & 14.4$\pm$3.5 & 12.9$\pm$1.3 & 13.6$\pm$2.3 & 19.7$\pm$2.6 & 84.8$\pm$1.3 & \textbf{87.9$\pm$1.3} & 84.8$\pm$1.3 & \textbf{86.4$\pm$2.3} \\
\bottomrule
\end{tabular}
\end{table}

CBGB (Section~\ref{sec:cbgb}) provides an external check on GRA, using geospatial problems authored outside this project with deterministic numeric answers. Both configurations with GRA's tools answer most questions correctly at every tier, and their scores decline moderately with the benchmark's difficulty tiers (Table~\ref{tab:cbgb-tiers}): GRA scores 95.6 and 97.8 on the Easy tier with Gemini 3.1 Pro and Gemini 3.8 Flash, 80.0 and 81.7 on the Intermediate tier, and 77.8 with both models on the Difficult tier. As on GeoKTools, upfront planning and execution guidance add no consistent gain on this benchmark  (Section~\ref{sec:res-main}).

Both baselines score far lower, between 12.9 and 19.7 overall, and fail almost all Difficult questions. They do solve a fraction of the Easy tier, with the standalone LLM at 24.4 and 22.2 and LLM + Basic Tools reaching 26.7 with Gemini 3.8 Flash. This is the pattern one expects of a benchmark that is still measuring something: a few questions are answerable from general knowledge, but most require retrieving and computing over satellite data.

As noted in Section~\ref{sec:cbgb}, our setup differs from the published CBGB protocol in several respects, so these numbers are not comparable with those reported in the original paper. The scores above use the benchmark's original criterion, a match to two significant figures.

Scored with the more permissive LLM-based autorater (Section~\ref{sec:geoktools}), the baselines gain 31--41 points while the tool-equipped configurations gain about 8, narrowing the gap without changing the ranking. This illustrates how proportional partial credit rewards plausible estimates computed from values recalled by the model or found on the web, rather than from the underlying data, and supports reporting the benchmark's original criterion.

\subsection{UnivEARTH: Benchmark Saturation}\label{sec:res-univearth}

\begin{table}[htbp]
\centering
\small
\setlength{\tabcolsep}{6pt}
\caption{UnivEARTH results (test set, 140 questions). Scores are 0--100, mean $\pm$ std over three runs.}
\label{tab:univearth}
\begin{tabular}{l *{2}{C{1.8cm}}}
\toprule
\textbf{Configuration} & \textbf{3.1P} & \textbf{3.8F} \\
\midrule
\textbf{LLM} & 86.9$\pm$1.8 & 93.6$\pm$0.7 \\
\textbf{LLM + Basic Tools} & 96.7$\pm$1.8 & 98.3$\pm$0.8 \\
\textbf{LLM + GRA Tools} & 97.9$\pm$0.7 & 98.8$\pm$0.4 \\
\textbf{GRA} & 89.5$\pm$0.4 & 98.1$\pm$0.4 \\
\bottomrule
\end{tabular}
\end{table}

The second external benchmark yields a different pattern. Every configuration scores between 86.9 and 98.8 (Table~\ref{tab:univearth}), against roughly 33\% for the best systems in the original evaluation \citep{kao2026univearth}. The standalone LLM, which has no means of accessing Earth Engine data at all, scores 86.9 with Gemini 3.1 Pro and 93.6 with Gemini 3.8 Flash, and LLM + Basic Tools matches or exceeds GRA with both models (96.7 and 98.3, against 89.5 and 98.1).

Inspection of the execution traces explains the contrast with CBGB (where the same standalone LLMs score 14.4 and 12.9). The standalone LLM answers 100\% of questions (and 90.2\% correctly) without performing any computation, and LLM + Basic Tools answers 95.6\% of questions without invoking code execution, relying instead on web search, and 5.7\% without invoking any tool at all (10.2\% with 3.1 Pro, 1.2\% with 3.8 Flash). The baselines appear simply to know or look up the answers, which is unsurprising given that the questions derive from published NASA Earth Observatory articles whose text states the findings. GRA, by contrast, loads its Earth Engine skill in 60.8\% of cases (59.0\% with 3.1 Pro, 62.6\% with 3.8 Flash), following the intended analytical pathway and incurring the execution failures (29.0\% of cases) that pathway exposes it to.

The consequence is that UnivEARTH assigns essentially the same score to a system that performs the required geospatial analysis and to one that never attempts it. We therefore report UnivEARTH as a finding about benchmark durability rather than about system capability.

\subsection{Model Comparison and Cost-Accuracy Tradeoffs}\label{sec:cost}

Our evaluation also compares two Gemini models across all configurations. Within GRA, Gemini 3.8 Flash outperforms Gemini 3.1 Pro on all four benchmarks: by 15.1 points on GeoExpertQueries (62.7 versus 47.6), 4.3 points on GeoKTools (85.9 versus 81.6), 8.6 points on UnivEARTH (98.1 versus 89.5), and 1.6 points on CBGB (86.4 versus 84.8), the last within run-to-run variation.

\begin{table}[htbp]
\centering
\footnotesize
\setlength{\tabcolsep}{2pt}
\caption{Estimated inference cost per query (USD).}
\label{tab:cost}
\begin{tabular}{l *{8}{C{1.5cm}}}
\toprule
\textbf{Benchmark} & \multicolumn{2}{c}{\textbf{LLM}} & \multicolumn{2}{c}{\begin{tabular}[b]{@{}c@{}}\textbf{LLM +} \\ \textbf{Basic Tools}\end{tabular}} & \multicolumn{2}{c}{\begin{tabular}[b]{@{}c@{}}\textbf{LLM +} \\ \textbf{GRA Tools}\end{tabular}} & \multicolumn{2}{c}{\textbf{GRA}} \\
\cmidrule(lr){2-3} \cmidrule(lr){4-5} \cmidrule(lr){6-7} \cmidrule(lr){8-9}
 & \textbf{3.1P} & \textbf{3.8F} & \textbf{3.1P} & \textbf{3.8F} & \textbf{3.1P} & \textbf{3.8F} & \textbf{3.1P} & \textbf{3.8F} \\
\midrule
\textbf{GeoExpertQueries} & \$0.028 & \$0.010 & \$0.115 & \$0.044 & \$0.512 & \$0.167 & \$0.688 & \$0.199 \\
\textbf{GeoKTools} & \$0.024 & \$0.012 & \$0.575 & \$0.112 & \$0.410 & \$0.147 & \$0.360 & \$0.122 \\
\textbf{CBGB} & \$0.025 & \$0.017 & \$0.514 & \$0.132 & \$0.140 & \$0.110 & \$0.151 & \$0.093 \\
\textbf{UnivEARTH} & \$0.006 & \$0.002 & \$0.029 & \$0.010 & \$0.116 & \$0.046 & \$0.271 & \$0.079 \\
\bottomrule
\end{tabular}
\end{table}

Table~\ref{tab:cost} reports the estimated LLM token inference cost per query (excluding external data and compute services such as Earth Engine), to be read alongside the scores in Table~\ref{tab:main-results}. The standalone LLM is by far the cheapest configuration, at under three cents per query, but it scores 0.0 on GeoExpertQueries and below 15 on CBGB, so the saving buys nothing usable.

For GRA, Gemini 3.8 Flash is cheaper than Gemini 3.1 Pro on all four benchmarks, by 38 to 71\%, while scoring higher on all of them, most clearly on GeoExpertQueries, at \$0.199 against \$0.688 with a 15.1-point advantage.

A less expected result is that GRA often costs less than LLM + Basic Tools while scoring far higher: on CBGB it is cheaper with both models, and with Gemini 3.1 Pro it costs \$0.151 against \$0.514 while scoring about six times higher (84.8 versus 13.6); with Gemini 3.1 Pro it is also cheaper on GeoKTools (\$0.360 against \$0.575). The baseline spends heavily on attempts that do not reach a correct answer, whereas GRA arrives at one by a more direct route.

With Gemini 3.8 Flash, upfront planning and execution guidance also add little cost: on GeoExpertQueries, GRA costs \$0.199 per query against \$0.167 for LLM + GRA Tools, so the 10.5-point gain comes at about three cents per query. GRA with Gemini 3.8 Flash is thus both the most accurate configuration on GeoExpertQueries and costs less than a third as much as GRA with Gemini 3.1 Pro.

\subsection{Capability and Tool Usage}\label{sec:res-tools}

\begin{table}[htbp]
\centering
\small
\setlength{\tabcolsep}{6pt}
\caption{Capability usage on GeoExpertQueries (test set, all three runs). Top: share of queries in which at least one skill from each capability pillar (Table~\ref{tab:eaia-pillars}) is loaded. Bottom: average activity per query. General-purpose Earth Engine skills are counted under Prediction \& Spatial Compute, since they serve all domains. Map-display skills are omitted.}
\label{tab:tool-usage}
\begin{tabular}{l *{4}{C{1.7cm}}}
\toprule
 & \multicolumn{2}{c}{\begin{tabular}[b]{@{}c@{}}\textbf{LLM +} \\ \textbf{GRA Tools}\end{tabular}} & \multicolumn{2}{c}{\textbf{GRA}} \\
\cmidrule(lr){2-3} \cmidrule(lr){4-5}
 & \textbf{3.1P} & \textbf{3.8F} & \textbf{3.1P} & \textbf{3.8F} \\
\midrule
\textbf{Imagery \& Earth Observation} & $<$1\% & 2\% & 4\% & 3\% \\
\textbf{Weather, Climate \& Hazards} & 18\% & 18\% & 23\% & 18\% \\
\textbf{Population, Health \& Socioeconomics} & 21\% & 22\% & 23\% & 20\% \\
\textbf{POIs, Boundaries \& Mobility} & 83\% & 57\% & 93\% & 89\% \\
\textbf{Prediction \& Spatial Compute} & 83\% & 75\% & 93\% & 99\% \\
\hspace{1em}of which Earth Engine skills & 49\% & 38\% & 55\% & 54\% \\
\midrule
\textbf{Skills loaded per query} & 4.1 & 3.7 & 4.7 & 4.7 \\
\textbf{Python calls per query} & 3.6 & 6.6 & 3.0 & 5.4 \\
\textbf{Web searches per query} & 2.6 & 7.0 & 1.2 & 3.2 \\
\bottomrule
\end{tabular}
\end{table}

Table~\ref{tab:tool-usage} shows how often LLM + GRA Tools and GRA use each capability pillar (Table~\ref{tab:eaia-pillars}) on GeoExpertQueries, measured as the share of queries in which at least one skill from the pillar is loaded (the full catalog is listed in Appendix~\ref{app:tools}). We use skill loads rather than raw tool calls, since many of the most frequent calls, such as updating plan steps, are bookkeeping rather than analysis. GRA uses prediction and spatial compute skills (scratch datasets, spatial processing and Earth Engine) and POIs, boundaries and mobility skills on the large majority of queries, with Earth Engine alone on about half. Weather, climate and hazards skills, and population, health and socioeconomics skills, are each loaded on fewer than a quarter of queries, reflecting the topical spread of the benchmark. Imagery and Earth observation skills are loaded on at most 4\% of queries, but this understates the use of imagery, since much of it runs through the general-purpose Earth Engine skills, which we count under prediction and spatial compute. The most specialized skills, such as weather forecasting and regional similarity search, are each loaded on under 3\% of queries, and targeted benchmarks for them are part of our future work (Section~\ref{sec:conclusions}).

GRA's upfront planning and execution guidance change how these capabilities are used. With Gemini 3.8 Flash, LLM + GRA Tools relies heavily on web search, at 7.0 searches per query; GRA halves this to 3.2 and uses Earth Engine on more queries (54\% versus 38\%), a shift toward its geospatial capabilities that accompanies GRA's 10.5-point gain over LLM + GRA Tools on GeoExpertQueries. With Gemini 3.1 Pro, the shift is smaller. Within GRA, Gemini 3.8 Flash also runs substantially more Python than Gemini 3.1 Pro (5.4 versus 3.0 calls per query), in line with its greater readiness to write custom code (Section~\ref{sec:res-GeoExpertQueries}); the extra calls are mostly successful ones rather than retries after failed attempts.

\section{Conclusions and Future Work}\label{sec:conclusions}

As LLM-powered geospatial systems move toward real-world deployment, understanding the value of domain-specific tooling becomes critical. We compared four configurations of the same model: a standalone LLM, the LLM with basic tools, the LLM with GRA's skills and tools but without its upfront planning and execution guidance, and GRA itself. The evaluation spans four benchmarks, covering expert-level open-ended analysis, controlled multi-tool orchestration, and community-developed geospatial problems with deterministic answers. Two of these benchmarks are new. GeoExpertQueries introduces a higher level of evaluation complexity, with expert-authored golden workflows that capture domain best practices independently of any specific tool set. GeoKTools checks final answers against reference solvers across the breadth of GRA's tool suite, and is designed to measure how performance varies with the number of required tool types. Together with two externally developed benchmarks, CBGB and UnivEARTH, this suite provides substantially broader coverage than prior evaluations of geospatial AI systems, most of which rely on a single benchmark or on detailed analysis of specific use cases. We release the development splits of both new benchmarks in full.

Our results show that GRA, combining Gemini with a suite of geospatial tools and agentic orchestration, provides a substantial and consistent advantage over baselines that lack these capabilities. Across GeoExpertQueries, GeoKTools, and CBGB, GRA significantly outperforms both the standalone LLM and the LLM augmented with basic tools. The contrast is starkest on GeoExpertQueries, where GRA carries out end-to-end analyses that the baselines cannot meaningfully approximate. Comparing the four configurations shows that GRA's geospatial skills and tools drive the largest gain, while upfront planning and execution guidance add a substantial further improvement on open-ended problems with Gemini 3.8 Flash, a reminder that orchestration must be matched to the model to make the most of the underlying tools. The advantage remains large on CBGB and GeoKTools, where baselines partially compensate through parametric knowledge but fall well short of tool-assisted performance, and on GeoKTools it holds for every number of required tool types. The exception is UnivEARTH, where all examined systems score above 85, including systems with no access to Earth Engine at all. Within GRA, the more recent Gemini 3.8 Flash outperforms Gemini 3.1 Pro on GeoExpertQueries and GeoKTools at substantially lower cost, making it the preferred model in these evaluations.

Our evaluation also surfaces a methodological finding: geospatial benchmarks decay in two distinct ways, and our two external benchmarks illustrate one each. UnivEARTH has become saturated, with a standalone LLM scoring 87 to 94, indicating that its answers are now largely recoverable from the widely published source articles from which its questions were derived. CBGB decays differently. Four of its reference values no longer matched the data at the time we ran it, because Sentinel-2 imagery had been reprocessed and a protected area boundary had been redrawn. Contamination is widely discussed in the evaluation literature, but drift in the underlying data is more specific to domains that measure the physical world, and it implies that geospatial benchmarks require periodic revalidation of their ground truth rather than one-time construction.

Taken together, these results indicate that pairing a strong general-purpose model with domain-specific geospatial tooling and agentic orchestration substantially extends what is practically answerable. Many questions that would otherwise require considerable expert effort, and that a general-purpose model alone cannot handle, can now be addressed in large part within a single interaction. Because the golden workflows specify steps rather than reference values, these results establish process alignment rather than outcome correctness; GRA's output is best seen as a first draft that experts can review and build on.

Looking ahead, our research will focus both on advancing the capability of geospatial reasoning agents and on improving the evaluation of their performance.

\subsection{Geospatial Reasoning Capability}

GRA incorporates capabilities such as the spatiotemporal prediction tools, which the current benchmarks do not exercise, and others, such as the Remote Sensing Foundations models and WeatherNext, which they exercise lightly (Section~\ref{sec:res-tools}). While companion work \citep{ma2026} demonstrates the quantitative gains of the Population Dynamics Foundation Model (PDFM) and AlphaEarth Foundations embeddings on supervised spatial prediction and downscaling tasks, we intend to develop targeted analytical benchmarks that isolate open-vocabulary visual search and ensemble weather forecasting (WeatherNext) in multi-step workflows. In addition, we are extending GRA's ability to discover and select relevant data across large catalogs, a central step in open-ended geospatial analysis.

We expect domain-specific tooling and orchestration to remain important as models improve. Geospatial models, data, and tools for reasoning and forecasting will keep advancing too, raising the bar for planetary intelligence. Knowledge about the planet is vast, detailed, and continually changing, as the data drift we observed in CBGB illustrates, and no fixed set of model parameters can hold it; the value of a strong model therefore lies increasingly in reasoning and in using the right tools, and modular skills let the system's capabilities grow without retraining the model.

\subsection{Evaluating Geospatial Reasoning Agents}

Considerable work remains in evaluation as much as in capability, and several limitations of our current evaluation approach deserve mention. Our results here cover Gemini models only; extending the comparison to other LLM families is left for future work. We exclude queries that cannot be completed, for example those requiring proprietary or unavailable data, so our results do not capture how systems behave when a task is genuinely infeasible. The LLM + Basic Tools baseline runs without network access from its code interpreter, for the reasons given in Section~\ref{sec:systems}; a network-enabled baseline, able to retrieve geospatial data directly from public sources, would be a useful complementary comparison, and we leave that configuration to future work. GeoKTools templates were designed around operations supported by GRA's tools, which favors the configurations that use them, so the benchmark is best read as a test of multi-tool orchestration rather than of general geospatial coverage. The comparison between GRA and LLM + GRA Tools measures the orchestration layer as a whole; it does not separate the planning stage from individual instructions, such as preferring dedicated data sources over web search or planning around substitutes in best-effort mode, and isolating their contributions is left for future work.

Our GeoExpertQueries scores come from an LLM autorater. Although Section~\ref{sec:ar-validation} validates the autorater with domain experts who rated the same traces relative to the golden workflows, absolute scores depend on both the validity of the workflow and the strictness of the judge, and a stricter autorater would likely assign lower scores across the board. Our conclusions therefore rest mainly on comparisons \emph{between} configurations, which we expect to hold.

Reliably rating open-ended geospatial analyses is an open problem that we hope the wider community will take up. Evaluating GRA on real-world enterprise queries would provide the most direct measure of practical utility, though this poses challenges around data availability, sensitivity and reproducibility. Producing high-quality geospatial benchmarks is expensive and time consuming, particularly for GeoExpertQueries-style evaluations that require multiple independent expert workflows per query. Scaling this process, for example by using dedicated agents to assist in benchmark construction, is an important direction that we intend to pursue. Finer-grained autoraters that can audit whether proxy dataset substitutions or changes in processing method are scientifically defensible also remain an open question for future work.

We see the evaluation presented here as a step in the right direction, in both its approach and its concrete benchmarks. Notably, on the open-ended problems that most closely resemble real geospatial practice, we currently observe a clear gap between the full GRA with Gemini 3.8 Flash and all other configurations examined. While this gap may change as models and methodologies evolve, it points to GRA's promise for helping answer complex questions about our planet.

\section*{Acknowledgments}

On behalf of the authors, we would like to express our heartfelt gratitude to Genady Beryozkin, Adam Boulanger, David Fork, Yotam Gigi, 
Jules Kuperminc, Tomer Shekel, and Natalie Williams for their valuable input and support. We are especially grateful to Simon Ilyushchenko for his thorough and critical review of our evaluation, which substantially improved this report.

\bibliography{main}

\appendix
\section{GRA Details}

\subsection{GRA Tool Suite}\label{app:tools}

Table~\ref{tab:capabilities} catalogs GRA's capabilities, grouped by the five pillars of Table~\ref{tab:eaia-pillars} plus cross-cutting functions, with the underlying data sources and models.

\begingroup
\footnotesize
\setlength{\tabcolsep}{3pt}
\begin{longtable}{>{\raggedright\arraybackslash}p{2.1cm}>{\raggedright\arraybackslash}p{2.9cm}>{\raggedright\arraybackslash}p{5.5cm}>{\raggedright\arraybackslash}p{4.7cm}}
\caption{GRA capabilities by pillar (* developed at Google).}\label{tab:capabilities}\\
\toprule
\textbf{Category} & \textbf{Capability} & \textbf{Analytical Functionality \& Operations} & \textbf{Underlying Data Source or Model} \\
\midrule
\endfirsthead
\caption[]{GRA capabilities by pillar (continued)}\\
\toprule
\textbf{Category} & \textbf{Capability} & \textbf{Analytical Functionality \& Operations} & \textbf{Underlying Data Source or Model} \\
\midrule
\endhead
\midrule
\multicolumn{4}{r}{\emph{Continued on next page}}\\
\endfoot
\bottomrule
\endlastfoot
Imagery \& Earth Observation & Embedding-based surface change detection & Annual surface-similarity maps and vectorized change polygons between two years, from cosine similarity of 64-dimensional satellite embeddings & AlphaEarth Foundations* \citep{brown2025alphaearth} (in Google Earth Engine* \citep{gorelick2017google}) \\
\rowrule
 & Terrain slope & Zonal slope statistics (mean, min, max percent grade) from a 30 m elevation model over administrative or custom geometries & Copernicus DEM GLO-30 \citep{esa2022copdem} \\
\rowrule
 & Open-vocabulary overhead imagery search \& classification & Zero-shot search for objects or scenes, and open-vocabulary land-cover/scene classification, over high-resolution US imagery (2023--2025) & Remote Sensing Foundations models* \citep{barzilai2025rsfm}, Google overhead imagery* \\
\midrule
Weather, Climate \& Hazards & Point weather forecasts & Current, hourly and daily point forecasts (temperature, precipitation, wind, humidity, UV, cloud cover) & Google Maps Data* \\
\rowrule
 & Ensemble forecast statistics & Regional ensemble mean, spread and percentiles & WeatherNext* \citep{google2025weather} \\
\rowrule
 & Tropical cyclones & Ensemble track, wind-field and strike-probability forecasts for active storms; historical storm tracks and wind-swath impact areas & Google cyclone model* (FNV3, via Weather Lab) \citep{alet2025cyclone}; NOAA IBTrACS v4 \citep{knapp2010ibtracs}; NOAA NHC strike probabilities \\
\rowrule
 & Flood forecasting \& streamflow reanalysis & Riverine and flash-flood events, forecast inundation polygons, gauge forecasts and metadata; daily streamflow reanalysis (1980--2023) and return-period thresholds & Google Flood Forecasting* \citep{nearing2024floods}; GRRR* \citep{nearing2024floods} \\
\rowrule
 & Floodplain screening & Flags geometries intersecting geomorphic floodplains, low-slope terrain or frequently inundated surface water & GFPLAIN250m \citep{nardi2019gfplain}; Copernicus DEM GLO-30 \citep{esa2022copdem}; JRC Global Surface Water \citep{pekel2016gsw} \\
\rowrule
 & Historical weather & Hourly to annual statistics, anomalies and threshold-exceedance areas over custom regions & ERA5 \citep{hersbach2020era5}, ERA5-Land \citep{munozsabater2021era5land}, NASA GPM IMERG V07 \citep{huffman2023imerg}, PRISM (4 km) \citep{daly2008prism} \\
\rowrule
 & Climate-hazard projections & Projections to 2100 under SSP2-4.5, SSP3-7.0 and SSP5-8.5 for extreme heat, precipitation, wildfire, riverine flooding and water stress, with composite risk assessment & NEX-GDDP-CMIP6 \citep{thrasher2022nexgddp}; WRI Aqueduct 4.0 \citep{kuzma2023aqueduct}; JRC GloFAS flood hazard \citep{baugh2024glofas}; USDA Wildfire Risk to Communities \citep{scott2024wrc} \\
\rowrule
 & Air quality \& pollen & Current, forecast and historical AQI, pollutant concentrations and health guidance; air-quality heatmaps; pollen forecasts & Google Maps Data* \\
\midrule
Population, Health \& Socioeconomics & Socioeconomic, demographic \& health statistics & Place resolution, statistical-variable discovery, and cross-sectional and time-series tables across administrative levels & Data Commons* \citep{guha2023datacommons} (e.g., US Census ACS, CDC, FEMA NRI, World Bank) \\
\rowrule
 & Regional similarity search & Most similar or dissimilar US counties or ZIP codes from learned embeddings & PDFM* \citep{agarwal2024pdfm} \\
\midrule
POIs, Boundaries \& Mobility & Administrative boundaries & Gazetteer lookup, polygon retrieval (countries to census tracts and ZIP codes), reverse geocoding and spatial filtering & Who's On First \citep{whosonfirst}; US Census TIGER/Line \& ZCTA \citep{uscensus2020tiger} \\
\rowrule
 & Points of interest & POI search and filtering within polygons, counts across areas, and representative centerpoints & Google Maps Data* \\
\rowrule
 & Routing \& accessibility & Routes, travel-time matrices and isochrones & Google Maps Data* \\
\rowrule
 & Rooftop solar potential & Building-level solar capacity, panel layout and savings; solar-flux, surface-elevation and shade rasters & Google Maps Data* \\
\midrule
Prediction \& Spatial Compute & Scratch datasets \& sandboxed Python & Geospatial data is kept outside the model's context window using dataset references; dataset inspection; sandboxed Python (e.g., pandas, GeoPandas, Shapely, NumPy, pyproj, Earth Engine API), custom Earth Engine scripts (spectral indices, cloud masking, compositing) & GRA scratch datasets \\
\rowrule
 & Geometric processing & CRS-aware buffering (automatic local projection), pairwise distances and spatial-intersection filtering & GRA scratch datasets \\
\rowrule
 & Earth Engine analysis (any dataset) & Server-side reduction (mean, median, min, max, sum) of any gridded catalog dataset (e.g., elevation, nightlights, temperature) over regions & Google Earth Engine* catalog \citep{gorelick2017google} (e.g., Sentinel-1/2, Landsat, MODIS, VIIRS, Sentinel-5P, ESA WorldCover, Dynamic World*, Open Buildings*, WorldPop) \\
\rowrule
 & Time-series forecasting & Multi-step forecasts of regional statistical series & TimesFM* \citep{das2024timesfm} with PDFM* embeddings \citep{agarwal2024pdfm} \\
\rowrule
 & Regional prediction \& super-resolution & Regression from population-dynamics embeddings to fill unobserved regions; county-to-ZIP downscaling & PDFM* \citep{agarwal2024pdfm} \\
\rowrule
 & Automated spatial modeling & Covariate curation, feature selection, spatial/panel regression, nowcasting and coarse-to-fine super-resolution & Planetary Prediction Engine \citep{ma2026} (PDFM*, AlphaEarth Foundations*, Data Commons*, TimesFM*) \\
\midrule
Cross-cutting & Upfront planning & Capability-aware multi-step plan before any geospatial tool call, with progress tracking during execution & GRA planner \\
\rowrule
 & Web search & Auxiliary lookups (event dates, dataset IDs, thresholds, reference values) & Grounding with Google Search* \\
\end{longtable}
\endgroup

\subsection{Progressive Disclosure and Prefix Caching}\label{app:progd}

Packaging tools behind progressively disclosed skills introduces a trade-off between tool-calling reliability and serving efficiency that is worth recording, since it applies to any agent adopting this pattern.

In our implementation, loading a skill registers that skill's native function declarations with the inference API's tool configuration. This preserves the model's post-training alignment for function calling and allows schema-validated decoding, which matters for the complex argument structures typical of geospatial tools. The cost is that inference APIs generally require tool declarations to sit in the request configuration with the system instruction, ahead of the conversation history. Mutating the tool configuration part-way through a trajectory therefore changes the request structure near its start and invalidates the serving key-value prefix cache past that point, forcing the accumulated conversation to be re-prefilled. Loading skills one at a time, after many steps or large tool outputs have accumulated, is correspondingly expensive.

Two mitigations exist. The first is protocol-level: OpenAI's Responses API introduced tool search and additional-tools items in March 2026,\footnote{OpenAI tool search documentation: \url{https://developers.openai.com/api/docs/guides/tools-tool-search\#add-tools-at-a-specific-point-in-the-input}} which allow appending newly loaded tool schemas to the conversation history rather than to the request configuration, preserving the cached prefix of earlier turns. This option is not yet available in the API used here. The second is architectural, and is how skill scripts work: an agent can keep its tool configuration static by exposing a single generic ``runner tool'' and invoking the desired functionality through arguments to that tool. This avoids cache invalidation on any API, but forgoes the dedicated function-calling training and schema validation that native declarations receive, which can lower invocation success rates.

GRA retains native function declarations for invocation reliability and mitigates the cache cost behaviorally, by instructing the agent to load the domain skills it expects to need early and together, before large tool outputs accumulate.

\subsection{LLM-based Autoraters}\label{app:autoraters}

Both autoraters are powered by Gemini 3.8 Flash with the default (medium) thinking level.

\subsubsection{GeoExpertQueries Workflow Autorater}

This section presents the system prompt used by the GeoExpertQueries autorater. Given a user query, the consensus golden workflow, and the agent's execution trace (including tool calls and their results), the autorater rates, on a scale from 0 to 2, how much of the golden workflow's analytical value the agent's executed work delivers. It treats the golden workflow as a reference rather than a checklist, and weighs deviations from it by their impact on the answer rather than by their number. When the golden workflow contains alternative expert approaches, the autorater rates the trace against the one it follows most closely. The autorater is implemented as a single Gemini call with the following system prompt:

\begin{lstlisting}
Today is {{ current_date }}. Even though this might seem to be in the future for you, keep in mind you are doing inference well past your training date.
You are an expert GIS analyst and geospatial data scientist.
Your task is to rate whether an AI agent carried out a valid analytical
workflow for a user's geospatial query. You compare the agent's trace with the
golden plan, an expert-written solution to the same query.
WHAT YOU RATE:
The golden plan shows what a geospatial expert would do: which data the answer
needs, and which analytical steps turn that data into the answer. Rate how much
of that analytical value the agent's executed work delivers: its tool calls,
the data they returned, and the processing applied to that data. A step the
agent only describes or promises is not executed.
- Do NOT judge whether the final numbers, arithmetic, or conclusions are
  correct, or penalize rounding or transcription errors in the final summary;
  rate the workflow steps and data sources the agent actually executed.
- Ignore the golden plan's Outputs section (maps, charts, formatted tables).
FOLLOWING THE GOLDEN PLAN:
The golden plan is a reference, not a checklist. Any approach counts as
following it if a geospatial expert would accept the approach as answering the
query equally well. Do not require the golden plan's exact datasets, tools,
platforms, or step order. A proxy for the data the query needs, or a simpler
method that answers a less precise question, is not an equivalent approach: it
is a deviation, weighed as described in HOW TO RATE.
If a step in the golden plan lists alternative methods or data sources, using
any one of them counts as following that step.
{%- if multiple_golden_workflows %}
This golden plan contains more than one workflow (Alternative 1,
Alternative 2, ...). Following any one of them counts as following the
golden plan. Rate against the workflow the trace follows most closely;
everywhere else in these instructions, "the golden plan" means that
workflow.
{%- endif %}
DATA SOURCES AND FALLBACKS:
- Real tool data vs hardcoded inputs: A step or data dimension counts as
  covered when its inputs come from tool outputs (retrieved datasets, queries,
  or search results, including values read off, summarized, or derived from
  them and passed into later code). Hardcoding empirical figures, observations,
  or coordinates from memory into a script does not cover that step.
- Tool outages and fallbacks: When a specialized tool fails and the agent
  pivots to a fallback using available tools, do not penalize the failed call
  itself. Rate the fallback like any other approach: it can earn Score 2 only
  if a geospatial expert would accept it as answering the query equally well;
  otherwise it is a deviation (a proxy or a shallower substitute), weighed as
  described in HOW TO RATE.
- Unspecified parameters: The agent MAY choose what the query leaves
  unspecified, such as study area bounds, time windows, thresholds, weights,
  scenario settings, or standard method constants. Never penalize reasonable
  choices.
HOW TO RATE (IMPACT, NOT STEP COUNT):
1. Identify each deviation from the golden plan: a skipped step, a shallower
   substitute, a different data source or method, narrower coverage, or a
   failed tool call without an adequate fallback.
2. Judge how much each deviation reduces the usefulness of the answer to a
   geospatial expert, compared with what the golden plan would deliver: none,
   slight, significant, or fatal.
3. Weigh deviations by their effect on the answer to the query's main question,
   not by how many there are. The deviations are fatal when the executed work
   no longer delivers what the golden plan's core steps deliver.
** CRITICAL PENALTY RULES **:
Each of the following means the agent's work is not a real solution and is an
AUTOMATIC FAIL (Score 0):
1. Plan only: the agent provides a plan or method description without
   executing it (unless the query explicitly asks for a plan).
2. No final answer: the run ends without one, for example because it stopped
   or timed out after intermediate steps.
3. No real analysis executed: the trace fails to execute any core data
   dimension of the golden plan on real tool-retrieved data (for example, it
   relies entirely on hardcoded/memory values or a generic web lookup where
   geospatial analysis is required), or fabricates tool outputs or observations.
DO NOT PENALIZE:
- Choices you would have made differently.
- Processing limits (a sample, a capped batch) that still serve the step's
  purpose.
- Skipping steps marked Optional.
- A different legitimate scoring or weighting scheme.
- Technical details such as projections, normalization, units, or resolution.
- Minor rounding or reordering of tool-produced values in the final answer, or
  overconfident wording.
- Missing maps, charts, or other visual deliverables.
INPUT DATA:
- User Query: the natural-language request.
- Golden Plan: the expert-verified solution (Datasets -> Steps -> Outputs).
- Agent Trace: the agent's messages, the tools it invoked, the results of those
  tool calls, and its final answer.
--- INPUTS ---
User Query:
<user_query>
{{ model_prompt }}
</user_query>
Golden Plan:
<golden_plan>
{{ golden_plan }}
</golden_plan>
Agent Trace:
<agent_trace>
{{ candidate_workflow }}
</agent_trace>
--- RATING SCALE ---
Rate the agent's work on a scale from 0 to 2.
Before scoring, write out:
{% if multiple_golden_workflows -%}
- which golden-plan workflow you rate against;
{% endif -%}
- the golden plan's core data dimensions and steps and, for each, whether the agent covered it with real tool outputs, used a fallback/proxy, or missed/hardcoded it;
- each deviation from the golden plan and its impact on usefulness (none, slight, significant, or fatal).
Keep in mind the ** CRITICAL PENALTY RULES ** listed above.
Score 0 (Not a real solution): A CRITICAL PENALTY RULE applies (plan only; no final answer; no real analysis executed), or the deviations are fatal: the executed work no longer delivers what the golden plan's core steps deliver.
Score 1 (Partial match): A real analysis ran on tool-produced data and addresses the query's main question, but deviations from the golden plan significantly reduce its usefulness to a geospatial expert. Examples: the trajectory covers only a subset of the core data dimensions in the golden plan with real tool data (while omitting others, replacing a core step with a much shallower substitute, or filling a missing dimension with estimates).
Score 2 (Full match): The executed work follows the golden plan, directly or through an approach (including a valid fallback) a geospatial expert would accept as equally good, covering the core data dimensions with no more than a slight loss of usefulness. These do not lower the score: simpler implementations that serve each step's purpose, reasonable choices for unspecified parameters, skipped Optional steps, and missing maps or charts.
Write your reasoning first. End your response with the score (0, 1, or 2).
\end{lstlisting}

\textbf{Note:} Per-item scores are computed by running this evaluation three times, mapping each score of 0, 1, or 2 to 0, 0.5, or 1.0, respectively, and taking the arithmetic mean, so each item scores between 0 and 1 (shown on a 0--100 scale in the results tables).

\subsubsection{GeoKTools Autorater}

This section presents the system prompt used by the GeoKTools autorater. Given a query, a golden ground-truth answer computed by the reference solver, and the system's response, the autorater assigns a continuous score between 0 and 1 reflecting how accurately the response matches the reference. The same autorater is also used to score CBGB answers with partial credit (Section~\ref{sec:cbgb}). The autorater is implemented as a single Gemini call with the following system prompt:

\begin{lstlisting}
You are a geospatial domain expert autorater. Your task is to evaluate whether the Model Response correctly answers the Query by comparing it against the Golden Ground Truth.
## Evaluation Process
1. **Understand the query's domain**: What kind of geospatial reasoning is required? (distance, entity lookup, weather/climate forecast, coordinate, count, spatial comparison, multi-part compound, etc.). The query determines what units and scale matter.
2. **Measure the gap in domain units**: How different is the Model Response from the Golden Ground Truth, expressed in the units that matter for this query? (km, mm of rain, °C, entity count, probability %). Treat the Golden Ground Truth as the authoritative reference.
3. **Classify answer accuracy**: Using the four-tier scale below, decide how a geospatial expert would perceive the magnitude of the difference:
   - **Accurate**: Model Response matches Golden Ground Truth within accepted domain precision and standard rounding. A geospatial expert would consider them equivalent.
   - **Minor Noise**: Technically different but the gap is negligible in real-world geospatial practice — would not change domain interpretation or decisions.
   - **Very Wrong**: Meaningfully deviates in a way that would mislead a practitioner or produce a materially different outcome.
   - **Severely Inaccurate**: Completely off — wrong primary entity, wrong sign/order of magnitude, threshold crossing failure, or uncomputed fallback where real computation was required.
4. **Assign a continuous score [0.0, 1.0]** proportionate to the domain gap:
   - 1.0 = exact match or equivalent within domain precision / standard decimal rounding.
   - Continuous scaling: Smoothly degrade the score from 1.0 down to 0.0 in direct proportion to the magnitude of the error relative to the reference scale.
   - Severe failure: Scores approach 0.0 when errors exceed acceptable domain tolerance or when fundamental entities/thresholds are missed.
## Essential Domain Rules
- **Authoritative Reference**: Treat the Golden Ground Truth as THE authoritative reference. Do NOT consult external real-world knowledge to validate either answer.
- **Floating Point & Rounding Tolerance**: Standard decimal rounding or formatting differences that preserve the underlying quantity within domain measurement precision should be treated as Accurate (score 1.0). Do not penalize standard floating-point rounding.
- **Proportional Error Scaling**: Continuous quantities should be scored proportionally based on relative accuracy rather than step functions. Do not give binary 0/1 scores to partially accurate continuous estimates unless the error is severe.
- **Multi-Element Arrays (Arithmetic Mean Rule)**: When responses contain a list/tuple of independent answers (e.g. multiple entities or properties), evaluate each element score independently in element_scores. The overall_score MUST be the exact arithmetic mean of element_scores.
- **Coupled Entity Invalidation**: When a sub-calculation depends on identifying a specific entity (e.g. finding a specific store then computing its distance), identifying the wrong entity invalidates subsequent dependent metrics (score 0.0 for that coupled pair).
- **Threshold & Occurrence Sensitivity**: When the reference is zero (0 occurrences, 0 days, 0 stations), predicting non-zero events represents a false positive event detection and should be penalized heavily (<= 0.10).
- **Formatting & Semantic Equivalence**: Differences in markdown code fences, JSON key order, whitespace, capitalization, trailing zeros, or standard state abbreviations with identical underlying value should be treated as Accurate (score 1.0).
- **Empty / Null Responses**: An empty or null model response when ground truth has content = Severely Inaccurate (0.0).
\end{lstlisting}

\textbf{Note:} Each call pairs this system prompt with a user message built from the following template, in which the placeholders in braces are filled with the query, the reference answer, and the system's response:

\begin{lstlisting}
Input Query:
{query}
Golden Ground Truth:
{golden_str}
Model Response:
{model_resp_str}
Please evaluate the Model Response against the Golden Ground Truth following the evaluation process, accuracy tiers, and essential domain rules. Provide your output in JSON format.
\end{lstlisting}

\section{GeoExpertQueries Examples Explained}\label{app:examples}

This appendix explains, in plain language, the three example queries in Table~\ref{tab:geoexpert-examples}: what each query means, why it matters, how the analysis laid out in its golden workflow proceeds, and what complicates it.

\subsection{Natural Hazards \& Crisis Resilience: 2022 Pakistan Floods Damage \& Reconstruction Prioritization}

\textbf{Prompt:} \emph{Using SAR Interferometry (InSAR) coherence change detection, estimate the structural damage in the hard-hit zones of the 2022 Pakistan floods in the Sindh and Balochistan provinces. Differentiate between inundated (water) and destroyed (rubble). Overlay with poverty maps to prioritize reconstruction grants.}

\begin{itemize}
  \item \textbf{What the question means:} During the catastrophic 2022 floods in Pakistan, clouds and weather blocked traditional satellite cameras. This task uses observations from \textbf{radar satellites}---which can ``see'' through clouds and at night---to compare the ground before and after the floods. Specifically, it separates places that are simply \textbf{underwater} (smooth water reflects radar signals away) from places where \textbf{buildings collapsed into rubble} (rough debris scatters radar signals and changes the surface structure between passes), and combines that damage map with local poverty data.
  \item \textbf{Why it's important:} Disaster relief budgets are limited, and field surveys take weeks or months when roads are washed out. Distinguishing standing water from collapsed homes---and cross-referencing with poverty maps---allows governments and aid agencies to rapidly direct reconstruction grants to the poorest communities that actually lost their homes.
  \item \textbf{How the Analysis Works:}
  \begin{itemize}
    \item \emph{Clean \& align radar images:} Correct for terrain distortions and radar noise across pre- and post-flood images.
    \item \emph{Separate water from rubble:} Filter analysis strictly to pre-existing towns. Flag areas where radar signal drops sharply as \textbf{standing floodwater}, and areas where surface stability (``coherence'') drops while radar bounce remains rough/high as \textbf{collapsed building rubble}.
    \item \emph{Rank need:} Combine the density of destroyed buildings with poverty and population levels for each local district (union council).
  \end{itemize}
  \item \textbf{Complications:}
  \begin{itemize}
    \item Structural damages to building exterior might be harder to establish using purely SAR -- it might require adding optical imagery taken from an angle or ground-level imagery.
  \end{itemize}
\end{itemize}

\subsection{Public Health: Emergency Obstetric Care Accessibility in Sierra Leone}

\textbf{Prompt:} \emph{Assess the physical accessibility of emergency obstetric care in Sierra Leone by calculating travel times over the Dynamic World land cover friction surface.}

\begin{itemize}
  \item \textbf{What the question means:} Instead of measuring simple straight-line (``as the crow flies'') distances to hospitals, this analysis calculates \textbf{how long it actually takes a pregnant woman to reach an emergency maternity hospital} from every location in Sierra Leone. It builds a ``friction map'' that accounts for driving speeds on different types of roads, walking speeds through forests or fields, steepness of hills, and rivers that can only be crossed at bridges or ferries. Dynamic World is one of a variety of datasets describing land cover that labels every pixel as trees, crops, built-up, water, etc.
  \item \textbf{Why it's important:} Sierra Leone has a high maternal mortality rate. Knowing exactly which communities live more than two hours away---and how many women live there---tells health planners where to build new clinics, station ambulances, or upgrade roads and bridges.
  \item \textbf{How the Analysis Works:}
  \begin{itemize}
    \item \emph{Build a travel-speed (``friction'') map:} Assign realistic driving speeds to roads and walking speeds to off-road terrain (slowing walking speeds on steep slopes and blocking travel across rivers except at bridges/ferries).
    \item \emph{Calculate fastest routes:} Compute the minimum travel time from every 100m$\times$100m grid cell in the country to the nearest qualifying maternity facility.
    \item \emph{Measure population gaps:} Apply the 2-hour (120-minute) WHO travel time benchmark and overlay population maps to count how many women of childbearing age live inside vs. outside the 2-hour reach in each district.
  \end{itemize}
  \item \textbf{Complications:}
  \begin{itemize}
    \item A more sophisticated analysis would look at areas that are at risk of being cut off during major storms and flood events (for example, by using historical flood datasets to understand the frequency of flooding, as well as climate models).
    \item The model also assumes a vehicle is available and that each listed hospital is actually staffed and stocked around the clock -- it measures physical, not effective, access.
  \end{itemize}
\end{itemize}

\subsection{Site Selection \& Urban Planning: Building Footprints vs. Available Land in Buenos Aires}

\textbf{Prompt:} \emph{For the city of Buenos Aires, determine the Building Footprint area versus available land area.}

\begin{itemize}
  \item \textbf{What the question means:} This question asks for a city-wide space audit of Buenos Aires: how much of the city's land is \textbf{already covered by buildings}, how much is \textbf{off-limits for construction} (such as roads, rivers, parks, and protected zones), and how much \textbf{truly buildable vacant land} remains.
  \item \textbf{Why it's important:} More than half of humanity already lives in cities and the share keeps rising -- Argentina is already $\sim$92\% urban. That growth puts pressure on housing, transport, water and sewer networks, and green space. For example, urban planners and developers need to know whether the city still has room to build outward on vacant lots or if it must grow upward. Such planning also prevents illegal or unsafe development in flood zones, wetlands, and protected green space.
  \item \textbf{How the Analysis Works:}
  \begin{itemize}
    \item \emph{Define and clean the building layer:} Decide what counts as a building (permanent vs. temporary structures -- hardest in informal settlements), remove duplicate/overlapping footprints.
    \item \emph{Carve out ``no-build'' zones:} Combine roads, railways, rivers (plus safety buffers), and protected parks into an exclusion mask of land that cannot be developed.
    \item \emph{Calculate remaining space:} Subtract both the existing building footprints and the ``no-build'' exclusion mask from the total city land area to find the true \textbf{available buildable land}, then calculate percentages across neighborhoods.
  \end{itemize}
  \item \textbf{Complications:}
  \begin{itemize}
    \item ``Buenos Aires'' must be defined first -- the Autonomous City is essentially built out, so the interesting vacant land lies in the surrounding metropolitan area, where cadastral data is patchier and often requires local government cooperation. The analysis measures physical vacancy only: zoning determines what may legally be built, and 2D footprints say nothing about building height, so under-used lots in high-density zones are invisible to it.
  \end{itemize}
\end{itemize}

\end{document}